\documentclass[11pt,letterpaper,logo]{thuair}

\usepackage[utf8]{inputenc} % allow utf-8 input
\usepackage[T1]{fontenc}    % use 8-bit T1 fonts
\usepackage{hyperref}       % hyperlinks
\usepackage{url}            % simple URL typesetting
\usepackage{booktabs}       % professional-quality tables
\usepackage{amsfonts}       % blackboard math symbols
\usepackage{nicefrac}       % compact symbols for 1/2, etc.
\usepackage{microtype}      % microtypography
\usepackage{xcolor}         % colors
\usepackage[utf8]{inputenc} % allow utf-8 input
\usepackage[T1]{fontenc}    % use 8-bit T1 fonts
\usepackage{hyperref}       % hyperlinks
\usepackage{url}            % simple URL typesetting
\usepackage{booktabs}       % professional-quality tables
\usepackage{amsfonts}       % blackboard math symbols
\usepackage{nicefrac}       % compact symbols for 1/2, etc.
\usepackage{microtype}      % microtypography
\usepackage{xcolor}         % colors
\usepackage{graphicx}
\usepackage{eso-pic}
\newcommand{\shijie}[1]{}

\usepackage{multirow}
\usepackage{amsmath}
\usepackage[table]{xcolor} % 加载支持表格颜色的宏包
\usepackage{colortbl}      % 可选，增强表格着色支持
\usepackage[ruled,vlined]{algorithm2e}
\usepackage{svg}

\usepackage{graphicx}
\usepackage{wrapfig}
\usepackage{subcaption}
\usepackage{threeparttable}
\usepackage{wrapfig} % 放在导言区
\usepackage{graphicx}
\usepackage{fontawesome5}

\usepackage{xcolor}
\newcommand{\up}[1]{\textcolor{green!60!black}{(+#1\%)}}
\newcommand{\down}[1]{\textcolor{red}{(-#1\%)}}
\newcommand{\base}[1]{\textcolor{gray}{($\pm$ #1\%)}}

\title{MemCompiler: Compile, Don't Inject — State-Conditioned Memory for Embodied Agents}

\author{%
  \textbf{Xin Ding}\textsuperscript{1}\hspace{4pt}\textsuperscript{$*$}
  \quad
\textbf{Xinrui Wang}\textsuperscript{2}\hspace{4pt}\textsuperscript{$*$}
  \quad
    \textbf{ Yifan Yang}\textsuperscript{3}
  \quad
  \textbf{ Hao Wu}\textsuperscript{4}\hspace{4pt}
  \quad
  \textbf{ Shiqi Jiang}\textsuperscript{3}
  \quad
  \textbf{ Qianxi Zhang}\textsuperscript{3}\\
  \quad
  \textbf{ Liang Mi}\textsuperscript{4}
  \quad
    \textbf{ Hanxin Zhu}\textsuperscript{1}
  \quad
      \textbf{ Kun Li}\textsuperscript{5}
  \quad
    \textbf{ Yunxin Liu}\textsuperscript{5}
  \quad
  \textbf{  Zhibo Chen}\textsuperscript{1} \textsuperscript{$\dag$}
  \quad
  \textbf{ Ting Cao}\textsuperscript{5}\hspace{4pt}\textsuperscript{$\dag$} \\
  \textsuperscript{1}University of Science and Technology of China 
  \quad
  \textsuperscript{2}Huazhong University of Science and Technology \\
  \quad
  \textsuperscript{3}Microsoft Research 
  \quad
  \textsuperscript{4} Nanjing University 
  \quad 
  \textsuperscript{5} Institute for AI Industry Research (AIR), Tsinghua University
  \\
}

\begin{document}

\begin{abstract}
Existing memory systems for embodied agents typically inject retrieved memory as static context at episode start, a paradigm we term \textit{Ahead-of-time Monolithic Memory Injection (AMMI)}. However, this static design quickly becomes misaligned with the agent’s evolving state and may degrade lightweight executors below the no-memory baseline. To address this, we propose \textbf{MemCompiler}, which reframes memory utilization as \textit{State-Conditioned Memory Compilation}. A learned Memory Compiler reads a structured \textit{Brief State} capturing the agent’s current execution state and dynamically selects and compiles only relevant memory into executable guidance. This guidance is delivered through a text channel and a latent \textit{Soft-Mem} channel that preserves perceptual information not expressible in text. Across AlfWorld, EmbodiedBench, and ScienceWorld, MemCompiler consistently improves over no-memory across open-source backbones (up to +129\%), matches or approaches frontier closed-source systems, and reduces per-step latency by ~60\%, demonstrating that state-aware memory compilation improves both effectiveness and efficiency. The code and data is available at \href{https://air-embodied-brain.github.io/MemCompiler/}{https://air-embodied-brain.github.io/MemCompiler/}.
\end{abstract}

\maketitle

% \begin{figure}[h]
% \centering
% \includegraphics[width=\textwidth]{figure/teaser_framework_compare}
% \caption{Two paradigms for memory utilization in embodied agents. (a)~AMMI injects the full task memory
%   $\mathcal{M}$ at episode start directly, the Executor must attend over all
%   of $\mathcal{M}$, burying valuable experience under irrelevant entries. (b)~SCMC (Ours) retains $\mathcal{M}$ as a source
%   library and compiles only state-relevant content $m^{*,t}$ at each step, ensuring the Executor receives precisely what
%   it needs.}
% \label{fig:overview}
% \end{figure}

\begin{figure}[h]
\centering
\includegraphics[width=0.9\textwidth]{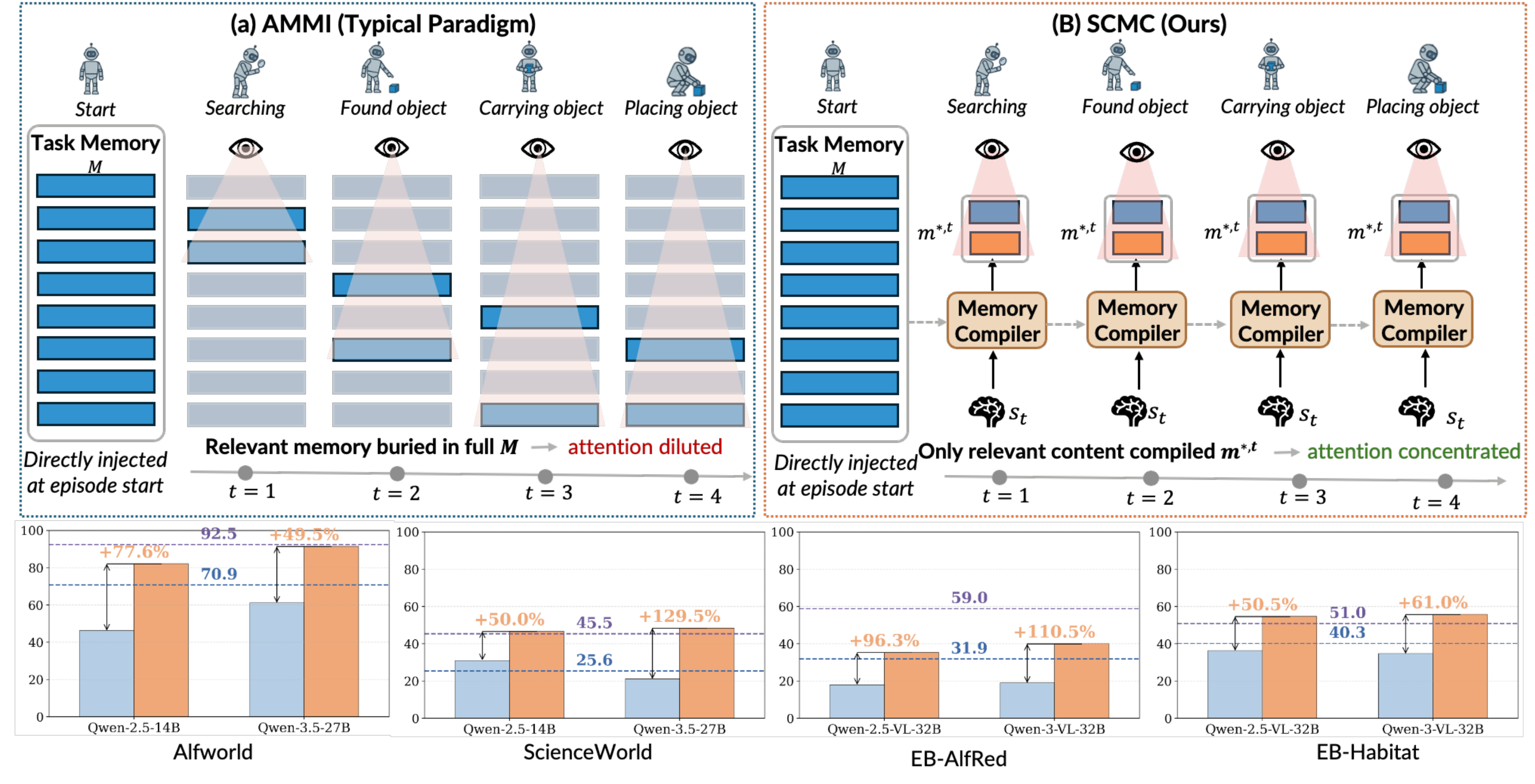}
\caption{Two paradigms for memory utilization in embodied agents. (a)~AMMI injects the full task memory
  $\mathcal{M}$ at episode start directly, the Executor must attend over all
  of $\mathcal{M}$, burying valuable experience under irrelevant entries. (b)~SCMC (Ours) retains $\mathcal{M}$ as a source
  library and compiles only state-relevant content $m^{*,t}$ at each step, ensuring the Executor receives precisely what
  it needs.
\includegraphics[height=0.5em]{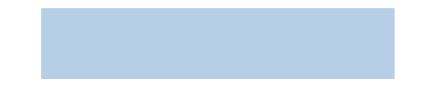},\includegraphics[height=0.2em]{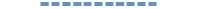},\includegraphics[height=0.2em]{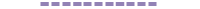},\includegraphics[height=0.5em]{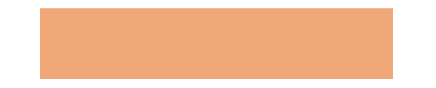} denote Qwen, GPT-5.2, Gemini-3-Flash, and our method, respectively, 
with performance measured across benchmarks.
}
\label{fig:overview}
\end{figure}

% Please add the following required packages to your document preamble:
% \usepackage{multirow}

% \begin{figure}[t]
% \centering
% \includegraphics[width=0.98\textwidth]{figure/memcompiler_framework_v2}
% \caption{Overview of MemCompiler at step $t$. The Memory Compiler reads runtime state $s_t = (o_t, b_t)$ and task memory $\mathcal{M}$ (retrieved once at episode start), then delivers the compiled result $m^{*,t}$ to the Executor through two parallel channels: a text channel ($m^{*,t}_{\text{text}}$) and a latent soft channel ($m^{*,t}_{\text{soft}}$), which are fused at the Executor's embedding level. Brief State $b_t$ is dynamically maintained via structured operations and fed back as part of the next step's state.}
% \label{fig:overview}
% \end{figure}

\section{Introduction}
\label{sec:intro}

For embodied agents deployed in persistent real-world environments such as homes and warehouses~\cite{agrawal2023physical,feng2026multi,team2025gemini,glocker2025llm}, memory is indispensable. It enables agents to overcome partial observability, accumulate cross-episode skill transfer, and avoid the long tail of rare failures that only accumulated experience can prevent~\cite{szot2021habitat,szot2023large,zheng2022vlmbench}. 
\vspace{-0.5mm}

However, current memory systems for embodiment have focused overwhelmingly on \textit{what} and \textit{how} to remember~\cite{cao2025remember,yan2025memory,ouyang2025reasoningbank,rezazadeh2024isolated}, and left largely unsolved the question of \textit{when} and \textit{how} to deliver memory during execution.
All existing embodiment memory systems share a common delivery paradigm we term Ahead-of-time Monolithic Memory Injection (AMMI)~\cite{limbacher2020h,chhikara2025mem0,wu2025sgmem,hu2025hiagent,long2025seeing}: retrieved entries are assembled into a static context at episode start and held fixed throughout execution. The static injection grows increasingly misaligned with current needs, burying valuable experience under irrelevant or even counterproductive entries, a failure mode we term \textit{attention dilution}. In our measurement, executor's attention over AMMI-injected memory tokens largely drops with episode steps 
%drops by 86.7\% by step 29 
(Figure~\ref{fig:attention_ana}). It may even lead to \textit{worse results} than no memory.
\vspace{-0.5mm}

% This attention dilution arises from structural properties specific to embodiment that are absent in language agents. In language agents, memory usage can often rely on the executor’s powerful ability to retrieve relevant information from a static text context. In contrast, embodied agents must infer memory relevance online from a continuously evolving multimodal execution state, a challenge further exacerbated by the fact that embodied policy models are typically lightweight compared to  large language models (LLMs). Concretely, this difficulty stems from two factors. First, memory relevance is state-dependent~\cite{shridhar2020alfred,li2023behavior,kim2025robot}: whether an entry is useful depends on the joint state of completed subgoals and current environmental beliefs, both of which must be inferred from a continuous visual-action stream rather than an explicit query. As a result, monolithic memory injection cannot adapt to shifting execution phases and may actively degrade performance over time. Second, embodied experience is inherently multimodal~\cite{team2025gemini,nasiriany2026robocasa365,srivastava2022behavior}: visual layouts, spatial relationships, and object affordances are difficult to faithfully capture in text, while task-relevant signals are often spatially sparse in visual inputs. This leads text-only memory to lose critical cues, whereas naive visual injection further amplifies attention dilution.

This attention dilution arises from structural properties specific to embodiment that are absent in language agents. In language agents, memory usage can often rely on the executor’s strong ability to retrieve relevant information from a static text context. In contrast, embodied agents must infer memory relevance online from a continuously evolving multimodal execution state, a challenge further amplified by the fact that embodied policy models are typically lightweight compared to language agents~\cite{openclaw2026,anthropic_claude_code}. Specifically, (i) memory relevance is \textbf{state-dependent}~\cite{shridhar2020alfred,li2023behavior,kim2025robot}, meaning its usefulness depends on inferred subgoal progress and environmental beliefs derived from a continuous visual-action stream rather than an explicit query, making static memory injection unable to adapt to state shifts and thereby degrading performance; and (ii) embodied experience is inherently \textbf{multimodal}~\cite{team2025gemini,nasiriany2026robocasa365,srivastava2022behavior}, where spatial and affordance information is difficult to encode in text while task-relevant signals are sparse in visual inputs, causing text-only memory to lose critical cues and naive visual images injection to further amplify attention dilution.
\vspace{-0.5mm}

To address both challenges, we propose \textbf{MemCompiler}, a new paradigm called \textbf{State-Conditioned Memory Compilation (SCMC)}. This method replaces AMMI’s monolithic pre-injection mechanism with a state-aware, multimodal-sensitive memory compilation process, ensuring that the guidance delivered to the Executor is always aligned with current execution needs. In classical compilation, a compiler transforms source code into target instructions conditioned on the runtime environment; the same source code yields different executables for different targets. We draw an analogy: task memory is source code, the agent's current execution state is runtime context, and a learned Memory Compiler dynamically translates relevant memory entries into actionable guidance at each decision step. 
\vspace{-0.5mm}

To realize MemCompiler, we introduce the Memory Compiler and Soft-Mem as two complementary components. The Memory Compiler, a lightweight model, consults the agent's current Brief State at each step to identify which entries in task memory are currently applicable and compile them into targeted text guidance. Leveraging the same internal representations, Soft-Mem further projects latent soft tokens directly into the Executor's embedding space, transmitting perceptual and spatial knowledge that the text channel cannot encode; an orthogonality constraint prevents the latent channel from collapsing into a copy of the text output. Both components are trained end-to-end via supervised fine-tuning, followed by GRPO-based~\cite{shao2024deepseekmath} reinforcement learning that sharpens the Memory Compiler's compilation decisions using binary task-success rewards.
\vspace{-0.5mm}

We evaluate MemCompiler on AlfWorld, EmbodiedBench (EB-ALFRED, EB-Habitat), and ScienceWorld across four open-source and two closed-source backbones. MemCompiler consistently improves over the no-memory baseline across all open-source executors and benchmarks, with gains reaching +110\% on EB-ALFRED and +129\% on ScienceWorld, while AMMI baselines often degrade smaller executors. Notably, MemCompiler enables mid-size open-source models to approach or match frontier closed-source systems, with Qwen-3.5-27B + MemCompiler matching Gemini-3-flash on AlfWorld and surpassing GPT-5.2 and Gemini-3-flash (no memory) on EB-Habitat. Beyond accuracy, MemCompiler is also more efficient: compared to AMMI on the same backbone, it reduces executor input tokens by ~60\% and cuts per-step latency from 0.30s to 0.12s, as the executor only processes compact, state-relevant compiled memory.

\begin{figure}[t]
\centering
\includegraphics[width=0.95\textwidth]{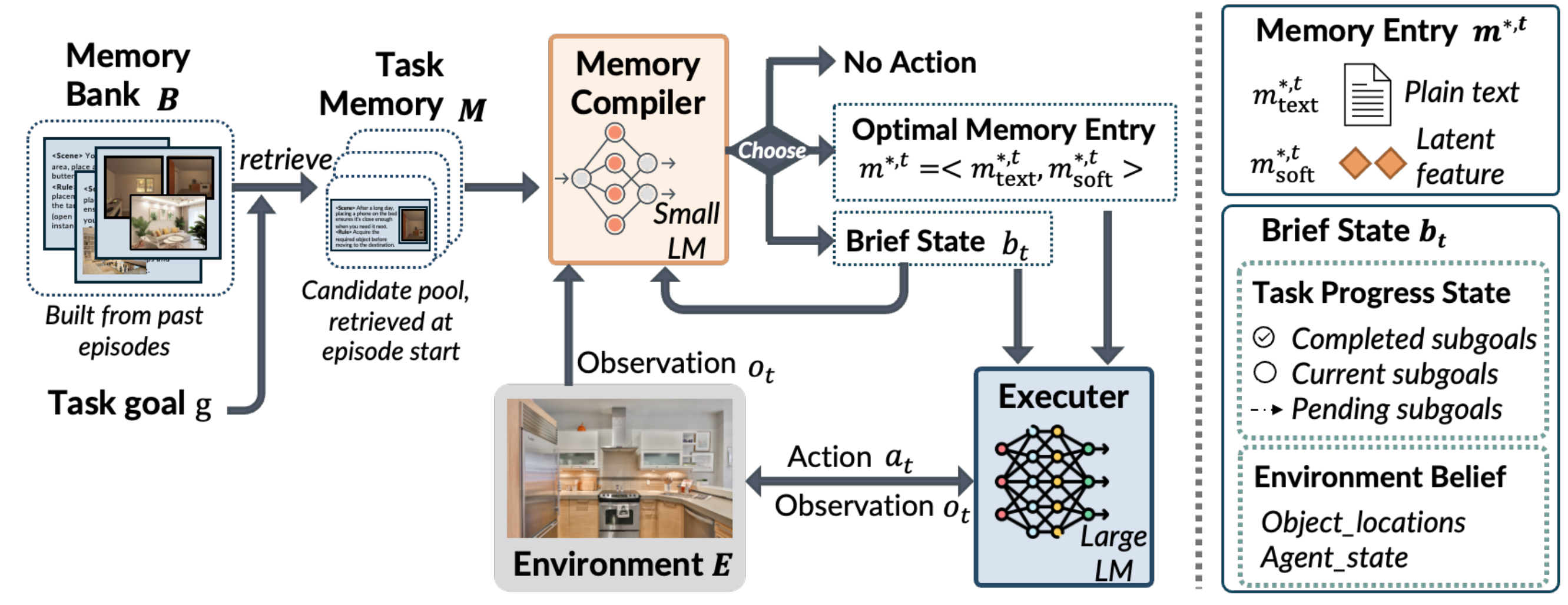}
\caption{Overview of MemCompiler at step $t$. The Memory Compiler reads runtime state $s_t = (o_t, b_t)$ and task memory $\mathcal{M}$ (retrieved once at episode start), then delivers the compiled result $m^{*,t}$ to the Executor through two parallel channels: a text channel ($m^{*,t}_{\text{text}}$) and a latent soft channel ($m^{*,t}_{\text{soft}}$), which are fused at the Executor's embedding level. Brief State $b_t$ is dynamically maintained via structured operations and fed back as part of the next step's state. Note, at each step the Memory Compiler decides among four output types (Section~\ref{sec:memcompiler}): \textsc{experience} ($m^{*,t}$ only), \textsc{brief} ($\Delta b_t$ only), \textsc{hybrid} (both jointly), or \textsc{noaction} (neither, when no entry in $\mathcal{M}$ is currently applicable).
}
\label{fig:overview}
\vspace{-5mm}
\end{figure}

\section{Related Work}
% As discussed in Section~\ref{sec:intro}, existing memory systems for embodied agents are limited along two fundamental axes: they assume static memory delivery that does not adapt to execution, and they rely primarily on textual representations that fail to capture perceptual experience.

\paragraph{General Agent Memory Systems.}
Most general-purpose agent memory systems assume a \emph{static, upfront memory injection} paradigm, where retrieved or summarized content is provided once and remains unchanged throughout execution. Foundational systems such as Generative Agents~\cite{park2023generative,wang2024mobile,team2025gemini}, Reflexion~\cite{shinn2023reflexion}, MemGPT~\cite{packer2023memgpt}, and ExpeL~\cite{zhao2024expel} store and retrieve experience in natural language, injecting it into the input context during execution. 

Subsequent work explores more structured memory organization, including knowledge graph indexing as in HippoRAG~\cite{gutierrez2024hipporag}, LLM-driven memory management as exemplified by Mem0~\cite{chhikara2025mem0}, Zettelkasten-style~\cite{kadavy2021digital,ahrens2022take} linking proposed in A-Mem~\cite{xu2025mem}, and agent-controlled hot-path updates introduced in LangMem. Code agents such as Claude Code~\cite{anthropic_claude_code} and OpenClaw~\cite{openclaw2026} introduce rule-based progressive disclosure, incrementally injecting references during execution. However, these systems are primarily developed for language- or code-centric agents built on powerful LLM backbones~\cite{anthropic2026claudeopus46,brown2020language,team2023gemini}, and thus assume text-based interaction and high-capacity reasoning, making them less suitable for embodied agents operating over visual-action streams under lightweight policy constraints.

\paragraph{Memory for Embodied Agents.}
In embodied settings, prior work largely follows an \emph{Ahead-of-time Monolithic Memory Injection (AMMI)} paradigm, where memory is retrieved at task initialization and remains fixed thereafter. Representative systems include Voyager~\cite{wang2023voyager}, JARVIS-1~\cite{wang2024jarvis}, G-Memory~\cite{zhang2025g}, and Evo-Memory~\cite{wei2025evo}, which differ in memory structure but share this static injection scheme. While effective in providing task-level context, such approaches cannot adapt memory usage to the agent’s evolving execution state or shifting informational needs over time.

A smaller line of work moves toward \emph{query-time or step-level memory construction}. GAM~\cite{yan2025general} defers context construction to query time, and RoboMemory~\cite{lei2025robomemory} updates spatial and episodic memory at each step in real-world robotic settings. However, these approaches primarily focus on improving retrieval and organization, rather than explicitly modeling fine-grained state-dependent usefulness of memories during execution. In contrast, our Memory Compiler maintains a structured Brief State that enables explicit, state-conditioned compilation at each step.

\paragraph{Beyond Text: Latent Memory Representations.}
% Text-based memory captures procedural knowledge and abstract reasoning, but fundamentally struggles to represent visual layouts, spatial configurations, and object affordances that are central to embodied experience.
Existing approaches explore latent memory along different directions, yet fall short of the requirements of embodied execution. NextMem~\cite{zhang2026nextmem} derives latent representations by compressing textual memory, inheriting the information bottleneck of text as the primary source. MemGen~\cite{zhang2025memgen} generates latent tokens from the current reasoning state, but does not explicitly leverage cross-episode experience. VisMem~\cite{yu2025vismem} introduces perceptual latent tokens for visual reasoning, yet treats them largely as standalone augmentations, without conditioning on a structured execution state or integrating them with accumulated experience. As a result, latent memory remains decoupled from both the agent’s evolving task state and its accumulated experience.

In contrast, within our state-conditioned compilation framework, Soft-Mem selectively compiles latent representations from cross-episode memory and explicitly constrains them to complement, rather than duplicate, the textual channel.

\section{Method}
\subsection{Problem Formulation of MemCompiler}
\label{sec:formulation}
The problem investigated in this work is how to enable an embodied agent to autonomously decide \textit{when} and \textit{how} to utilize cross-episode experiential memory. The agent operates in environment $\mathcal{E}$ with state space $\mathcal{S}$ and action space $\mathcal{A}$, receiving visual observation $o_t \in \mathbb{R}^{H \times W \times C}$ and executing action $a_t \in \mathcal{A}$ at each timestep $t$ to pursue task goal $g$, where $s_t \in \mathcal{S}$ denotes the agent's full episode state. The agent has access to a memory bank $\mathcal{B}$ accumulated across prior episodes, storing strategies, failure reflections, and procedural guidelines. At the start of each episode, relevant experience entries are retrieved from $\mathcal{B}$ to form a fixed task memory candidate pool:  
\begin{equation}
  \mathcal{M} = \text{Retrieve}(\mathcal{B},\ g)
  \end{equation}
\paragraph{Ahead-of-Time Memory Injection.} We consider a standard memory utilization setting: Ahead-of-time Monolithic Memory Injection (AMMI), where $\mathcal{M}$ is directly injected into the agent's context at episode start, yielding a sequence of actions $a_{0:T}$ under policy $\pi(a_t \mid o_t,\ \mathcal{M})$ throughout execution.

However, the optimal memory entry $m^{*,t}$ is state-dependent:
  \begin{equation}
  m^{*,t} = \underset{m \in \mathcal{M}}{\arg\max}\ \mathbb{E}_{\tau \sim \pi(\cdot \mid s_t,
  m)}\bigl[\mathbf{1}(\text{success}(\tau))\bigr]
  \end{equation}
As execution proceeds, subgoals are progressively completed and environmental observations accumulate, causing the agent's state to continuously evolve. Yet AMMI injects the full $\mathcal{M}$ at episode start without subsequent selection. As the applicability of each entry in $\mathcal{M}$ shifts with the evolving state, the Executor is forced to attend over the entire static injection, with valuable experience increasingly buried under irrelevant or even counterproductive entries.

\paragraph{State-Conditioned Memory Compilation.} To address this dilution, we propose State-Conditioned Memory Compilation (SCMC). We define \textbf{Brief State} $b_t = \phi(g,\ a_{0:t-1},\ o_{0:t-1})$ as a runtime context, where $\phi$ denotes a structured, task-relevant compression of the task goal $g$, action history $a_{0:t}$, and prior observations $o_{0:t-1}$, maintained and updated by the \textbf{Memory Compiler} throughout the episode. Defining the full \textit{runtime state} as $s_t = (o_t,\ b_t)$, the Memory Compiler takes the current $s_t$ and candidate pool $\mathcal{M}$ as
input at each timestep $t$, and dynamically \textbf{compiles} state-appropriate memory:
\begin{equation}
   m^{*,t} = \pi_C(s_t,\ \mathcal{M})
   \label{equ:m* generation}
\end{equation}
Under this formulation, $\mathcal{M}$ serves as \textit{source code}, and the Memory Compiler compiles $m^{*,t}$ conditioned on the current state $s_t$ at each step. The complete design of MemCompiler is detailed in Section~\ref{sec:memcompiler}.

\subsection{Methodology of MemCompiler}
\label{sec:memcompiler}
\subsubsection{Overview and Motivation}
MemCompiler addresses state-conditioned memory utilization for embodied agents. As established in Section~\ref{sec:formulation}, AMMI injects the full $\mathcal{M}$ at episode start without subsequent selection, burying valuable experience under irrelevant or counterproductive entries as $s_t$ evolves.

Resolving this requires answers to two questions. \textbf{When to compile:} memory compilation should be conditioned on the agent's current runtime state, requiring a compilation policy that decides at each step whether to compile, and a structured runtime context that captures task progress and environmental knowledge to support such decisions precise. \textbf{How to compile:} even given correct timing, compiled memory must be delivered while preserving information critical for decision-making, yet experiential memory $\mathcal{M}$ contains visual and spatial knowledge from prior episodes that text output alone cannot fully convey. These two questions motivate the design of MemCompiler, illustrated in Figure~\ref{fig:overview}.

\subsubsection{When to Compile}

\paragraph{Limitation of Static Injection.}
The fundamental limitation of AMMI lies in its inability to adapt memory usage over time. Injecting the full memory set $\mathcal{M}$ at episode start, before any task progress or environmental observations are available, results in a persistent mismatch between injected content and the agent's evolving state. As execution proceeds, relevant entries become increasingly obscured by irrelevant ones, leading to cumulative attention dilution.

\begin{figure}[t]
\centering
\includegraphics[width=0.9\textwidth]{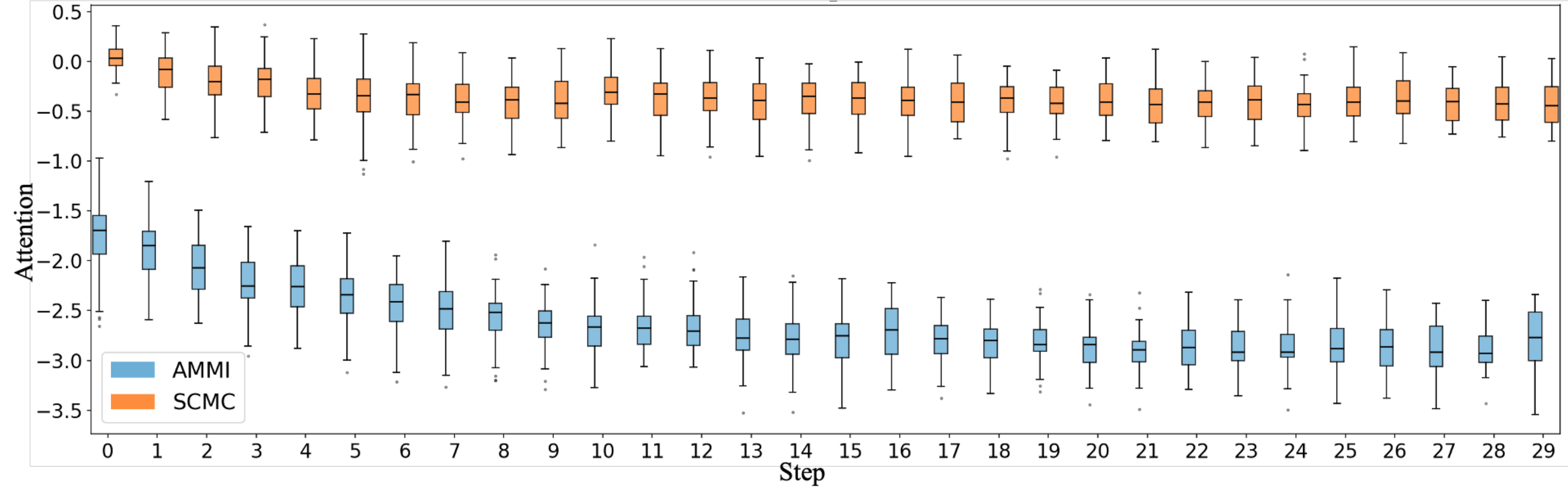}
\caption{Mean pre-softmax attention logit (higher values mean more attention weight) of the Executor over memory tokens across AlfWorld episodes (steps 0--29).  AMMI~(blue) exhibits monotonic decay, reflecting increasing misalignment between static memory and the agent's evolving state. In contrast, SCMC~(orange) maintains stable high attention throughout, indicating consistent alignment between memory and the Executor's current needs.}
\label{fig:attention_ana}
\vspace{-5mm}
\end{figure}
\paragraph{Empirical Evidence.}
We validate this on AlfWorld by visualizing the Executor's attention over memory tokens at each step (Figure~\ref{fig:attention_ana}). Under AMMI, attention declines monotonically throughout the episode, dropping by 86.7\% by step 29, as static injection becomes increasingly misaligned with the agent's evolving state. In contrast, under SCMC, attention remains stable across all steps, indicating that compiled memory stays consistently aligned with the Executor's current needs. This gap widens over time, confirming that attention dilution is an accumulating cost of static injection.
\vspace{-1mm}

\paragraph{Why Rule-based Disclosure Is Insufficient.}
A natural alternative is rule-based progressive disclosure, as used in code agents \cite{openclaw2026,anthropic_claude_code}, where memory is injected according to predefined trigger rules. This approach assumes that the applicability of a memory entry can be approximated by simple, observable conditions.
\vspace{-1mm}

However, in embodied settings, applicability depends on the joint state of task progress and environment belief, both of which evolve continuously and are not directly observable from the raw action-observation stream. As a result, rule-based reference are inherently misaligned with true utility. Moreover, unlike frontier language models, the lightweight policy executor lacks the capacity to filter large injected contexts, making incorrect injections costly in terms of attention.
\vspace{-1mm}

\paragraph{Brief State.}
At any timestep $t$, assessing whether a memory entry $m$ is applicable requires answering two questions: (1) does $m$'s guidance target the agent's \textit{current subgoal}? (2) do $m$'s assumed conditions hold in the \textit{current environment}? 
  
However, raw observation history cannot answer these questions directly: it is long, redundant, and lacks explicit task progress indices or organized environmental facts. We formalize the runtime state as $s_t = (o_t,\ b_t)$, where \textit{Brief State} $b_t = \phi(g,\ a_{0:t-1},\ o_{0:t-1})$ provides exactly these two signals through a two-dimensional structure:
\begin{itemize}
  \item \textbf{Task Progress State}: goal $g$, completed subgoals, current
  subgoal, and pending subgoals — answering \textit{whether $m$'s guidance
  is relevant to the current subgoal}.
  \item \textbf{Environment Belief}: known object locations, agent status,
  and task-relevant environmental facts — answering \textit{whether $m$'s
  assumed conditions currently hold}.
\end{itemize}
Together, these two dimensions provide the structured context necessary for step-level utility assessment.

\paragraph{Memory Compiler.}
  Given the structured runtime state $s_t = (o_t,\ b_t)$, the Memory Compiler faces two orthogonal decisions at each step: (1) should it provide memory guidance to the Executor? (2) should it update the Brief State for use in subsequent steps? 
  
  While Equation~\ref{equ:m* generation} captures the core case where $\pi_C$ directly outputs $m^{*,t}$, the full output space of $\pi_C$ accounts for both decisions, yielding four output types:
  \begin{itemize}
      \item \textbf{EXPERIENCE}: the compiled memory $m^{*,t}$ delivered as
      strategic guidance to the Executor, without updating Brief State.
      \item \textbf{BRIEF}: Brief State update $\Delta b_t$ only,
      refining the runtime context without guiding the Executor.
      \item \textbf{HYBRID}: both $m^{*,t}$ and $\Delta b_{t}$ simultaneously.
      \item \textbf{NOACTION}: $\emptyset$; no entry in $\mathcal{M}$ reaches
      sufficient utility given the current $s_t$, and the Executor then acts without memory guidance.
  \end{itemize}
  These four cases define the complete output space of $\pi_C$:
  \begin{equation}
      \pi_C(s_t,\ \mathcal{M}) \in
      \bigl\{m^{*,t},\ \Delta b_t,\ (m^{*,t},\ \Delta b_t),\ \emptyset\bigr\}
  \end{equation}
When the output includes $\Delta b_t$, Brief State updates follow $b_{t+1} = \text{Apply}(\Delta b_t,\ b_t)$, where $\Delta b_t$ specifies one of four operations: \textsc{Create} (new belief entry), \textsc{Update} (revised belief), \textsc{Delete} (invalidated belief), or \textsc{Fold} (belief state transition on subgoal completion).

\subsubsection{How to Compile}
\paragraph{Overview} In embodied tasks, experiential memory accumulated across episodes inherently contains perceptual information such as visual layouts and spatial configurations. These signals are critical for the agent's decisions yet resist faithful expression in natural language. We therefore introduce \textit{Soft-Mem}, a dual-channel compilation mechanism that delivers the compiled result $m^{*,t}$ to the Executor in both textual $m^{*,t}_{\text{text}}$ and latent $m^{*,t}_{\text{soft}}$ form.

However, training the latent channel solely through end-to-end action prediction loss leaves the relationship between the two channels unconstrained: since $m^{*,t}_{\text{soft}}$ has no ground-truth supervision during SFT, there is no guarantee it encodes information distinct from the text channel, leaving the RL stage with a poorly initialized latent channel. We therefore introduce $\mathcal{L}_{\text{soft}}$ to explicitly shape the latent channel before RL training begins.

\paragraph{Dual-Channel Output.} After autoregressively generating $m^{*,t}_{\text{text}}$ in text form, the Memory Compiler continues to process $N$ appended special \texttt{[UNK]} tokens. The final-layer hidden states at these positions, $\mathbf{h} \in \mathbb{R}^{N\times
d_{\text{MC}}}$, have attended over the full content of $\mathcal{M}$ including its visual and spatial information. A trainable projection $f_\phi$ maps $\mathbf{h}$ to the parameters of a Gaussian distribution in the Executor's embedding space:
  \begin{equation}
      (\boldsymbol{\mu}, \log\boldsymbol{\sigma}) = f_\phi(\mathbf{h}), \quad
       m^{*,t}_{\text{soft}} = \boldsymbol{\mu} + \boldsymbol{\sigma} \odot \boldsymbol{\epsilon}, \quad
      \boldsymbol{\epsilon} \sim \mathcal{N}(\mathbf{0}, \mathbf{I})
  \end{equation}
where $m^{*,t}_{\text{soft}} \in \mathbb{R}^{N \times d_{\text{base}}}$ then replaces the designated placeholder positions in the Executor's input embeddings, delivering latent perceptual semantics that bypass the text bottleneck. The compiled result received by the Executor thus consists of two parts: the textual $m^{*,t}_{\text{text}}$ carrying interpretable strategic guidance, and $m^{*,t}_{\text{soft}}$ carrying what text cannot say.

\paragraph{Training.}
We train MemCompiler in two stages. The \textbf{SFT stage} jointly optimizes three objectives: the text loss
$\mathcal{L}_{\text{text}}$ supervises the Memory Compiler's plaintext output via next-token prediction; the action loss $\mathcal{L}_{\text{action}}$ supervises the Executor's action prediction with $m^{*,t}_{\text{soft}}$ injected, with gradients flowing back through $m^{*,t}_{\text{soft}}$ to train $\mathbf{W}_p$; and the soft-token loss $\mathcal{L}_{\text{soft}}$ jointly constrains the information quality of $m^{*,t}_{\text{soft}}$:
\begin{equation}
      \mathcal{L}_{\text{soft}} =
      \underbrace{-\frac{1}{N}\sum_{i=1}^{N}\log p_\theta\!\left(
      \tilde{e}_i\right)}_{\text{semantic alignment}}
      + \lambda\,\underbrace{\left|\cos\!\left(\bar{h}_{\text{soft}},\
      \text{sg}[\bar{h}_{\text{text}}]\right)\right|}_{\text{complementarity constraint}}
  \end{equation}
where the semantic alignment term grounds the latent channel in perceptual content from $\mathcal{M}$ that text cannot express, while the complementarity constraint penalizes the cosine similarity between $\bar{h}_{\text{soft}}$ and the stop-gradient $\bar{h}_{\text{text}}$, preventing $m^{*,t}_{\text{soft}}$ from collapsing into a redundant copy of the text signal. Together, they act as an \textit{initialization prior} that simultaneously aligns the latent channel with perceptual content and pushes it away from the text representation before RL begins. The two terms are balanced by $\lambda$. The complete SFT objective is:
  \begin{equation}
      \mathcal{L}_{\text{SFT}} = \mathcal{L}_{\text{text}}
      + \mathcal{L}_{\text{action}}
      + \lambda_{\text{soft}}\,\mathcal{L}_{\text{soft}}
  \end{equation}
The \textbf{RL stage} applies GRPO~\cite{shao2024deepseekmath} to further refine the Memory Compiler's compilation policy, keeping the Executor fixed. For each episode, GRPO samples $G$ compilation trajectories from the current policy and assigns a binary reward $r_i \in \{0, 1\}$ based on task success:
  \begin{equation}
      \mathcal{L}_{\text{GRPO}} = -\frac{1}{G}\sum_{i=1}^{G}
      \min\!\left(
          \frac{\pi_\theta(o_i|s)}{\pi_{\theta_{\text{old}}}(o_i|s)} A_i,\
          \text{clip}\!\left(
              \frac{\pi_\theta(o_i|s)}{\pi_{\theta_{\text{old}}}(o_i|s)},
              1{-}\epsilon, 1{+}\epsilon
          \right) A_i
      \right)
  \end{equation}
where $A_i = \frac{r_i - \text{mean}(r_{1:G})}{\text{std}(r_{1:G})}$ is the group-normalized advantage. This stage enables the Memory Compiler to explore compilation strategies beyond the SFT distribution and exploit those that maximize task success.

% \vspace{300mm}

\section{Experiment}

\subsection{Experimental Setup}
\paragraph{Benchmarks.} We evaluate on three benchmarks spanning household manipulation, multimodal embodied planning, and scientific reasoning. AlfWorld~\cite{shridhar2020alfworld} provides 134 unseen household tasks across six categories (pick-and-place, examine-in-light, clean, heat, cool, pick-two-and-place) in 120 rooms, evaluated by task success rate. EmbodiedBench~\cite{yang2025embodiedbench} provides 1,128 tasks across two environments (EB-ALFRED, EB-Habitat), reporting success rate along six fine-grained capability dimensions: base ability (Base), commonsense reasoning (Com.), complex instruction following (Comp.), visual recognition (Vis.), spatial reasoning (Spat.), and long-horizon planning (Long). ScienceWorld~\cite{wang2022scienceworld} contains 30 interactive science experiments across 10 topics, spanning 7,200 parametric variations.

% we report task completion rate and partial completion score .

% \paragraph{Baselines.} We compare against four memory augmentation baselines spanning graph-structured and retrieval-based approaches: \textbf{G-Mem}~\cite{zhang2025g} and \textbf{A-Mem}~\cite{xu2025mem} organize memory as structured graphs with hierarchical or linked-note representations; \textbf{Mem0}~\cite{chhikara2025mem0} and \textbf{LangMem}~\cite{langchain2026langmem} maintain flat memory stores with LLM-driven extraction and vector retrieval. Despite differing in memory organization, all four share the AMMI injection paradigm: retrieved content is injected as static upfront context without step-wise adaptation. We additionally report \textbf{No Mem} as the base agent without memory augmentation. We evaluate across two closed-source backbones (GPT-5.2, Gemini-3-flash) and four open-source backbones: Qwen-2.5-14B, Qwen3.5-27B, Qwen-2.5-VL-32B and Qwen-3-VL-32B~\cite{bai2025qwen25vltechnicalreport,bai2025qwen3}.
\paragraph{Baselines.} We compare against five memory augmentation baselines spanning graph-structured, retrieval-based, and latent-token approaches: \textbf{G-Mem}~\cite{zhang2025g} and \textbf{A-Mem}~\cite{xu2025mem} organize memory as structured graphs with hierarchical or linked-note representations; \textbf{Mem0}~\cite{chhikara2025mem0} and \textbf{LangMem}~\cite{langchain2026langmem} maintain flat memory stores with LLM-driven extraction and vector retrieval; \textbf{MemGen}~\cite{zhang2025memgen} generates latent memory tokens from the executor's current reasoning state. The first four share the AMMI injection paradigm: retrieved content is injected as static upfront context without step-wise adaptation. We additionally report \textbf{No Mem} as the base agent without memory augmentation. We evaluate across two closed-source backbones (GPT-5.2, Gemini-3-flash) and four open-source backbones: Qwen-2.5-14B, Qwen3.5-27B, Qwen-2.5-VL-32B and Qwen-3-VL-32B~\cite{bai2025qwen25vltechnicalreport,bai2025qwen3}. 

To isolate the effect of memory delivery from memory curation, we keep the memory bank $\mathcal{M}$ deliberately minimal: each entry consists of a raw successful trajectory paired with  rationale explaining why it succeeded, without further distillation, scoring, or hierarchical reorganization.

\subsection{Main Results}

Tables~\ref{table1} and \ref{table2} report task performance across EmbodiedBench, AlfWorld, and ScienceWorld. Two patterns emerge.

\begin{table}[h]\fontsize{9}{10}\selectfont %字号，行间距
\caption{Success Rate (\%) on EmbodiedBench (EB-ALFRED and EB-Habitat) (\%). Capability dimensions: Base (base ability), Com.\ (commonsense reasoning), Comp.\ (complex instruction following), Vis.\ (visual recognition), Spat.\ (spatial reasoning), Long (long-horizon planning).}
% \vspace{-3mm}
\label{table1}
\renewcommand{\tabcolsep}{2pt} % 将列间距设为 3pt
\begin{tabular}{cclcccccclcccccc}
\toprule
\multirow{2}{*}{Executor}                                                      & \multicolumn{1}{c|}{\multirow{2}{*}{Method}} & \multicolumn{7}{c|}{EB-ALFRED}                                                                                                                                      & \multicolumn{7}{c}{EB-Habitat}                                                                                                                                 \\ \cline{3-16} 
& \multicolumn{1}{c|}{}                        & \multicolumn{1}{c}{Avg.}                   & Base                 & Com.               & Comp.              & Vis.               & Spat.              & \multicolumn{1}{c|}{Long} & \multicolumn{1}{c}{Avg.}                  & Base                 & Com.               & Comp.              & Vis.               & Spat.              & Long                 \\ \hline
\multicolumn{16}{c}{Close-Source MLLMs}      \\ \hline
\multirow{3}{*}{GPT-5}                                                      & \multicolumn{1}{c|}{No Mem}                  &    32  {\scriptsize \base{0}}                &          60            &         40            &    28                 &      36               &     28                 & \multicolumn{1}{c|}{0}     &     40.33 {\scriptsize \base{0}}                &     80                &          38            &   36                   &        40              &     32                &         16            \\
& \multicolumn{1}{c|}{Mem0}                    &        43.15 {\scriptsize \up{35.3}}             &         48             &        58             &     60                 &     47                &       36               & \multicolumn{1}{c|}{10}     &        37.67  {\scriptsize \down{6.6}}            &        76              &        36              &       30               &        44             &       30              &          10            \\
& \multicolumn{1}{c|}{G-Mem}                   &    45.33 {\scriptsize \up{42.1}}                 &          46           &      58               &       60               &       42               &         42             & \multicolumn{1}{c|}{24}     &       40.33 {\scriptsize \base{0}}              &          82            &          36            &           20           &         56            &        30             &        18             \\
 & \multicolumn{1}{c|}{A-Mem}                    &        46.33   {\scriptsize \up{45.2}}          &             52         &      54                &         48             &       34               &       42              & \multicolumn{1}{c|}{48}     &       32.33 {\scriptsize \down{19.8}}              &        64            &         30             &     18                &          48           &         28            &          6           \\
  & \multicolumn{1}{c|}{LangMem}                    &        46.67  {\scriptsize \up{46.3}}            &         74             &    56                &            62        &        46             &      38               & \multicolumn{1}{c|}{4}     &           43.33 {\scriptsize \up{7.4}}        &         84          &       40              &       40              &           56         &      34            &   6                  \\
                                                                            \hline
\multirow{3}{*}{Gemini}                                                     & \multicolumn{1}{c|}{No Mem}                  &   59.00 {\scriptsize \base{0}}                &           70          &        56            &    60                &        68             &     56                 & \multicolumn{1}{c|}{44}     & 51 {\scriptsize \base{0}}                &        86            &      52             &           46          &      62               &            30       &       30             \\
& \multicolumn{1}{c|}{Mem0}                    & \multicolumn{1}{l}{77.67{\scriptsize \up{31.7}}} & \multicolumn{1}{c}{82} & \multicolumn{1}{c}{80} & \multicolumn{1}{c}{82} & \multicolumn{1}{c}{80} & \multicolumn{1}{c}{74} & \multicolumn{1}{c|}{68}     & \multicolumn{1}{l}{57 {\scriptsize \up{11.8}}} & \multicolumn{1}{c}{86} & \multicolumn{1}{c}{ 64} & \multicolumn{1}{c}{48} & \multicolumn{1}{c}{70} & \multicolumn{1}{c}{30} & \multicolumn{1}{c}{44} \\
& \multicolumn{1}{c|}{G-Mem}                   &      79.27 {\scriptsize \up{34.4}}               &           82         &    82                  &      80                &     78                 &          76            & \multicolumn{1}{c|}{78}     &    68 {\scriptsize \up{33.3}}                 &      92                &        68              &  50                    &74                      &78                      &46                      \\ 
 & \multicolumn{1}{c|}{A-Mem}                    &      82.00  {\scriptsize \up{39.0}}              &         90             &      78                &        84              &      86                &           82           & \multicolumn{1}{c|}{72}     &    67.33  {\scriptsize \up{32.0}}                &    98                  &    72                  & 52                     &         78             &   80                   &         24             \\
  & \multicolumn{1}{c|}{LangMem}                    &     78.00  {\scriptsize \up{32.2}}               &       82               &76                      &78                      &84                      &76                      & \multicolumn{1}{c|}{72}     &62.67 {\scriptsize \up{22.9}}                     &               86       &            68          &     56                 &  78                    &74                      &14                      \\\hline
\multicolumn{16}{c}{Open-Source MLLMs}                                                                                                                                      \\ \hline
\multirow{7}{*}{\begin{tabular}[c]{@{}c@{}}Qwen-2.5\\ -VL-32B\end{tabular}} & \multicolumn{1}{c|}{No Mem}      &             18 {\scriptsize \base{0}}        &          22            &          26            &         24             &   12                   &       22               & \multicolumn{1}{c|}{2}     &        36.33 {\scriptsize \base{0}}             &      78                &   22                   &      52                &     26                 &              30        &        10              \\
& \multicolumn{1}{c|}{G-Mem}                   & \multicolumn{1}{l}{7.67 {\scriptsize \down{57.4}}} & \multicolumn{1}{c}{4} & \multicolumn{1}{c}{14} & \multicolumn{1}{c}{6} & \multicolumn{1}{c}{16} & \multicolumn{1}{c}{6} & \multicolumn{1}{c|}{0}     & \multicolumn{1}{l}{14.67 {\scriptsize \down{59.6}}} & \multicolumn{1}{c}{38} & \multicolumn{1}{c}{12} & \multicolumn{1}{c}{8} & \multicolumn{1}{c}{8} & \multicolumn{1}{c}{20} & \multicolumn{1}{c}{2} \\
& \multicolumn{1}{c|}{Mem0}                    & \multicolumn{1}{l}{14.66 {\scriptsize \down{18.6}}} & \multicolumn{1}{c}{20} & \multicolumn{1}{c}{18} & \multicolumn{1}{c}{26} & \multicolumn{1}{c}{12} & \multicolumn{1}{c}{10} & \multicolumn{1}{c|}{2}     & \multicolumn{1}{l}{38.33 {\scriptsize \up{5.5}}} & \multicolumn{1}{c}{68} & \multicolumn{1}{c}{34} & \multicolumn{1}{c}{44} & \multicolumn{1}{c}{38} & \multicolumn{1}{c}{30} & \multicolumn{1}{c}{16} \\
& \multicolumn{1}{c|}{A-Mem}                   & \multicolumn{1}{l}{3 {\scriptsize \down{83.3}}} & \multicolumn{1}{c}{2} & \multicolumn{1}{c}{4} & \multicolumn{1}{c}{4} & \multicolumn{1}{c}{6} & \multicolumn{1}{c}{2} & \multicolumn{1}{c|}{0}     & \multicolumn{1}{l}{8.33 {\scriptsize \down{77.1}}} & \multicolumn{1}{c}{24} & \multicolumn{1}{c}{8} & \multicolumn{1}{c}{6} & \multicolumn{1}{c}{8} & \multicolumn{1}{c}{4} & \multicolumn{1}{c}{0} \\
& \multicolumn{1}{c|}{Langmem}                    & \multicolumn{1}{l}{14.33 {\scriptsize \down{20.4}}} & \multicolumn{1}{c}{22} & \multicolumn{1}{c}{20} & \multicolumn{1}{c}{26} & \multicolumn{1}{c}{6} & \multicolumn{1}{c}{12} & \multicolumn{1}{c|}{0}     & \multicolumn{1}{l}{38.67 {\scriptsize \up{6.9}}} & \multicolumn{1}{c}{76} & \multicolumn{1}{c}{28} & \multicolumn{1}{c}{44} & \multicolumn{1}{c}{40} & \multicolumn{1}{c}{28} & \multicolumn{1}{c}{16} \\
& \multicolumn{1}{c|}{MemGen}                    & \multicolumn{1}{l}{14.33 {\scriptsize \down{20.4}}} & \multicolumn{1}{c}{14} & \multicolumn{1}{c}{10} & \multicolumn{1}{c}{16} & \multicolumn{1}{c}{6} & \multicolumn{1}{c}{8} & \multicolumn{1}{c|}{0}     & \multicolumn{1}{l}{21 {\scriptsize \down{42.2}}} & \multicolumn{1}{c}{54} & \multicolumn{1}{c}{8} & \multicolumn{1}{c}{18} & \multicolumn{1}{c}{18} & \multicolumn{1}{c}{24} & \multicolumn{1}{c}{4} \\

& \multicolumn{1}{c|}{Ours}                    &    \textbf{35.33 {\scriptsize \up{96.2}}}               &\textbf{42  }                     &\textbf{50   }                    &\textbf{44    }                   &\textbf{38 }                      &\textbf{34     }                  & \multicolumn{1}{c|}{4}     &\textbf{54.67  {\scriptsize \up{50.5}}   }                   &\textbf{84   }                &             \textbf{  54 }       &              \textbf{  56  }     &\textbf{60   }                   &\textbf{36        }               &\textbf{38  }                    \\ \hline
\multirow{7}{*}{\begin{tabular}[c]{@{}c@{}}Qwen-3\\-VL-32B\end{tabular}}   & \multicolumn{1}{c|}{No Mem}                  &        19  {\scriptsize \base{0}}            &        24             &       24               &       28               &         24             &      14                & \multicolumn{1}{c|}{0}     &          34.67  {\scriptsize \base{0}}          &         78             &    26                  &           28           &           34           &           28           &            14          \\
& \multicolumn{1}{c|}{G-Mem}                   & \multicolumn{1}{l}{25.33 {\scriptsize \up{33.3}}} & \multicolumn{1}{c}{34} & \multicolumn{1}{c}{36} & \multicolumn{1}{c}{38} & \multicolumn{1}{c}{24} & \multicolumn{1}{c}{14} & \multicolumn{1}{c|}{6}     & \multicolumn{1}{l}{44 {\scriptsize \up{26.9}}} & \multicolumn{1}{c}{88} & \multicolumn{1}{c}{24} & \multicolumn{1}{c}{36} & \multicolumn{1}{c}{50} & \multicolumn{1}{c}{32} & \multicolumn{1}{c}{34} \\
& \multicolumn{1}{c|}{Mem0}                    & \multicolumn{1}{l}{30.33 {\scriptsize \up{59.6}}} & \multicolumn{1}{c}{40} & \multicolumn{1}{c}{42} & \multicolumn{1}{c}{40} & \multicolumn{1}{c}{26} & \multicolumn{1}{c}{26} & \multicolumn{1}{c|}{8}     & \multicolumn{1}{l}{38.33 {\scriptsize \up{10.6}}} & \multicolumn{1}{c}{84} & \multicolumn{1}{c}{24} & \multicolumn{1}{c}{26} & \multicolumn{1}{c}{42} & \multicolumn{1}{c}{34} & \multicolumn{1}{c}{20} \\
& \multicolumn{1}{c|}{A-Mem}                   & \multicolumn{1}{c}{32.33 {\scriptsize \up{70.2}}} & \multicolumn{1}{c}{46} & \multicolumn{1}{c}{50} & \multicolumn{1}{c}{34} & \multicolumn{1}{c}{30} & \multicolumn{1}{c}{30} & \multicolumn{1}{c|}{4}     & \multicolumn{1}{c}{36.67 {\scriptsize \up{5.8}}} & \multicolumn{1}{c}{86} & \multicolumn{1}{c}{12} & \multicolumn{1}{c}{38} & \multicolumn{1}{c}{24} & \multicolumn{1}{c}{36} & \multicolumn{1}{c}{24} \\
& \multicolumn{1}{c|}{Langmem}                    & \multicolumn{1}{l}{31.33 {\scriptsize \up{64.9}}} & \multicolumn{1}{c}{38} & \multicolumn{1}{c}{44} & \multicolumn{1}{c}{44} & \multicolumn{1}{c}{28} & \multicolumn{1}{c}{26} & \multicolumn{1}{c|}{8}     & \multicolumn{1}{l}{38.33 {\scriptsize \up{10.6}}} & \multicolumn{1}{c}{86} & \multicolumn{1}{c}{26} & \multicolumn{1}{c}{28} & \multicolumn{1}{c}{40} & \multicolumn{1}{c}{32} & \multicolumn{1}{c}{18} \\
& \multicolumn{1}{c|}{MemGen}                    & \multicolumn{1}{l}{14.22 {\scriptsize \down{25.2}}} & \multicolumn{1}{c}{22} & \multicolumn{1}{c}{18} & \multicolumn{1}{c}{16} & \multicolumn{1}{c}{6} & \multicolumn{1}{c}{20} & \multicolumn{1}{c|}{4}     & \multicolumn{1}{c}{25.67 {\scriptsize \down{25.9}}} & \multicolumn{1}{c}{50} & \multicolumn{1}{c}{14} & \multicolumn{1}{c}{22} & \multicolumn{1}{c}{36} & \multicolumn{1}{c}{22} & \multicolumn{1}{c}{10} \\
& \multicolumn{1}{c|}{Ours}                    &  \textbf{  40  {\scriptsize \up{110}}          }     &\textbf{52   }                   &\textbf{54     }                 &\textbf{48    }                  &\textbf{38         }             &\textbf{38  }                    & \multicolumn{1}{c|}{\textbf{10}  }   &   \textbf{ 55.67 {\scriptsize \up{61.0}}         }         &90                  &\textbf{58      }                &       \textbf{  44  }       &\textbf{64        }              &\textbf{40   }                   &     \textbf{   38   }    \\ \bottomrule
\end{tabular}
\end{table}  
\begin{wraptable}{r}{0.4\linewidth}
% \vspace{-10pt}
\fontsize{7.5}{10}\selectfont
\caption{Success Rate on AlfWorld and ScienceWorld (\%).}
\label{table2}
\renewcommand{\tabcolsep}{3.5pt}
\begin{tabular}{llcc}
\toprule
\multicolumn{2}{c}{Executor / Method} & AlfWorld & ScienceWorld \\
\midrule
\multicolumn{4}{l}{\textit{Closed-Source MLLMs}} \\

\multirow{5}{*}{GPT-5.2} 
& No Mem & 70.89 {\scriptsize \base{0}}   & 25.55 {\scriptsize \base{0}}    \\
& G-Mem & 86.56 {\scriptsize \up{22.1}} & 51.51 {\scriptsize \up{101.6}} \\
& Mem-0 & 76.12 {\scriptsize \up{7.4}} & 22.22 {\scriptsize \down{13.0}} \\
& LangMem & 88.72 {\scriptsize \up{25.2}} & 46.67 {\scriptsize \up{82.7}} \\
& A-Mem & 92.54 {\scriptsize \up{30.5}} & 50.00 {\scriptsize \up{95.7}} \\

\cmidrule(l){2-4}

\multirow{5}{*}{\begin{tabular}[c]{@{}c@{}}Gemini-3\\-flash\end{tabular}} 
& No Mem & 92.54 {\scriptsize \base{0}}   & 45.55 {\scriptsize \base{0}}   \\
& G-Mem & 90.30 {\scriptsize \down{2.4}} & 44.44 {\scriptsize \down{2.4}} \\
& Mem-0 & 91.04 {\scriptsize \down{1.6}} & 42.86 {\scriptsize \down{5.9}} \\
& LangMem & 92.54 {\scriptsize \up{0.0}} & 50.82 {\scriptsize \up{}} \\
& A-Mem & 97.01 {\scriptsize \up{4.8}} & 47.78 {\scriptsize \up{4.9}} \\

\midrule
\multicolumn{4}{l}{\textit{Open-Source MLLMs}} \\

\multirow{6}{*}{\begin{tabular}[c]{@{}c@{}}Qwen-2.5\\-14B\end{tabular}} 
& No Mem & 46.27 {\scriptsize \base{0}}   & 31.11 {\scriptsize \base{0}}    \\
& G-Mem & 67.16 {\scriptsize \up{45.1}} & 32.22 {\scriptsize \up{3.6}} \\
& Mem-0 & 12.69 {\scriptsize \down{72.6}} & 27.78 {\scriptsize \down{10.7}} \\
& LangMem & 36.57 {\scriptsize \down{21.0}} & 15.56 {\scriptsize \down{50.0}} \\
& A-Mem & 11.19 {\scriptsize \down{75.8}} & 34.44 {\scriptsize \up{10.7}} \\
& MemGen & 42.54 {\scriptsize \down{8.1}} & 24.65 {\scriptsize \down{26.2}} \\
& \textbf{Ours} & \textbf{82.16} {\scriptsize \up{77.6}} & \textbf{46.67} {\scriptsize \up{50.0}}\\

\cmidrule(l){2-4}

\multirow{6}{*}{\begin{tabular}[c]{@{}c@{}}Qwen-3.5\\-27B\end{tabular}} 
& No Mem & 61.19 {\scriptsize \base{0}}   & 21.11 {\scriptsize \base{0}}   \\
& G-Mem & 74.62 {\scriptsize \up{22.0}} & 32.22 {\scriptsize \up{52.6}} \\
& Mem-0 & 62.69 {\scriptsize \up{2.5}} & 26.67 {\scriptsize \up{26.3}} \\
& LangMem & 62.69 {\scriptsize \up{2.5}} & 26.67 {\scriptsize \up{26.3}} \\
& A-Mem & 88.81 {\scriptsize \up{45.1}} & 32.22 {\scriptsize \up{52.6}} \\
& MemGen & 41.04 {\scriptsize \down{32.9}} & 14.32 {\scriptsize \down{32.2}} \\
& \textbf{Ours} & \textbf{91.45} {\scriptsize \up{49.5}} & \textbf{48.44} {\scriptsize \up{129}} \\
\bottomrule
\end{tabular}
\vspace{-25pt}
\end{wraptable}
\paragraph{AMMI baselines often hurt small executors; MemCompiler consistently helps.}
On open-source backbones, AMMI-style baselines frequently \emph{degrade} performance rather than improve it: across the four Qwen executors and four benchmarks, AMMI baselines yield negative gains in over half of all (executor, dataset) pairs, with extreme drops such as A-Mem at -83.3\% on Qwen-2.5-VL-32B / EB-ALFRED. The latent-token baseline MemGen exhibits similar degradation, indicating that latent representations alone, without state-conditioned compilation, do not suffice. In contrast, MemCompiler is the only method that improves over No Mem on every open-source backbone and benchmark, with relative gains reaching +110\% on EB-ALFRED and +129\% on ScienceWorld.

\paragraph{MemCompiler narrows the open-to-closed-source gap.}
With MemCompiler, mid-size open-source backbones close most of the gap to frontier closed-source executors. For example, Qwen-3.5-27B + Ours reaches 91.45\% on AlfWorld, essentially matching Gemini-3-flash No Mem (92.54\%); on EB-Habitat, Qwen-3-VL-32B + Ours (55.83\%) exceeds both GPT-5 and Gemini-3 No Mem. This indicates that the bottleneck for small open-source executors lies less in raw capacity than in how memory is delivered.

\section{Ablation}

\paragraph{Component Ablation.}
\begin{wraptable}{l}{0.5\linewidth}
\fontsize{7.5}{10}\selectfont
\vspace{-5mm}
  \caption{Component ablation. AlfWorld: success rate (\%). ScienceWorld: completion/partial score.}
  \label{tab:abl1}
  \begin{tabular}{lcc}
  \toprule
  Variant & AlfWorld & ScienceWorld \\
  \midrule
  MemCompiler (Full)                        & \textbf{82.16} & \textbf{46.67/69.33} \\
  \quad w/o Soft-Mem (text only)            & 78.93  & 42.22/65.15  \\
  \quad w/o Brief State (raw obs as state)  &73.94  & 38.50/60.67 \\
  AMMI (static injection, same memory)      & 35.03  & 27.93/52.68  \\
  No Memory                                 & 46.27 & 31.11/45.26 \\
  \bottomrule
  \end{tabular}
\vspace{-10pt}
\end{wraptable}
We ablate MemCompiler on AlfWorld and ScienceWorld with Qwen-2.5-14B, removing or replacing major components to isolate their contributions (Table~\ref{tab:abl1}): (i)~\textit{w/o Soft-Mem} removes the latent channel, passing only text guidance to the Executor; (ii)~\textit{w/o Brief State} replaces the structured two-dimensional belief state with the raw concatenation of goal and action history; (iii)~\textit{AMMI} injects the same retrieved $\mathcal{M}$ as static upfront context, isolating the contribution of dynamic compilation. 

Results show that removing either SCMC (via AMMI) or Brief State leads to the largest performance degradation, confirming that state-conditioned compilation is the core contributor. Soft-Mem provides additional gains particularly on ScienceWorld, where procedural perceptual knowledge benefits the latent channel.

\paragraph{Soft-Mem Design.}

We examine three design choices of Soft-Mem on EmbodiedBench EB-Habitat, focusing on Spat.\ and Vis.\ dimensions where latent perceptual information matters most (Table~\ref{tab:abl2}): (i) the number of soft tokens $N$; (ii) the orthogonality constraint; (iii) the projection type (Gaussian vs.\ linear). 

\begin{wraptable}{l}{0.5\linewidth}  % r=右侧，宽度=半栏
\vspace{-10pt} % 可调，控制上下间距
\fontsize{8}{10}\selectfont
\centering
\captionsetup{justification=centering}
  \caption{Soft-Mem ablation on EB-Habitat.}
  \label{tab:abl2}
\centering
\renewcommand{\tabcolsep}{4pt} % 稍微缩紧一点
 \begin{tabular}{lccc}
  \toprule
  Variant & Avg. & Spat. & Vis. \\
  \midrule
  $N=0$ (text only)                    &47.11  & 32 & 42 \\
  $N=4$                                & 51.09 & 28 & 50 \\
  $N=8$                                & 52.83 & 34 & 56 \\
  $N=16$ (default)                   & \textbf{54.67} & \textbf{36} & \textbf{60}  \\
  $N=32$                               & 53.17 & 36 & 50 \\
  \midrule
  $N=16$, w/o orth. constraint         & 50.33 & 32 & 52 \\
  $N=16$, linear proj. (no Gaussian)   & 46.23 & 30 & 48 \\
  \bottomrule
  \end{tabular}
% \vspace{-10pt}
\end{wraptable}

Results show that performance peaks at $N{=}16$ and degrades at larger $N$, suggesting diminishing returns from excess capacity.  Removing the orthogonality constraint causes the sharpest decline on Spat.\ and Vis., where text cannot recover latent perceptual information, confirming that without it the latent channel collapses toward the text representation. Replacing the Gaussian projection with a linear layer impedes the continued reinforcement learning process, leading to a moderate degradation in performance.
\begin{wrapfigure}{r}{0.4\textwidth} % r表示右侧，宽度可调
    \centering
\vspace{-10pt}
    \includegraphics[width=0.4\textwidth]{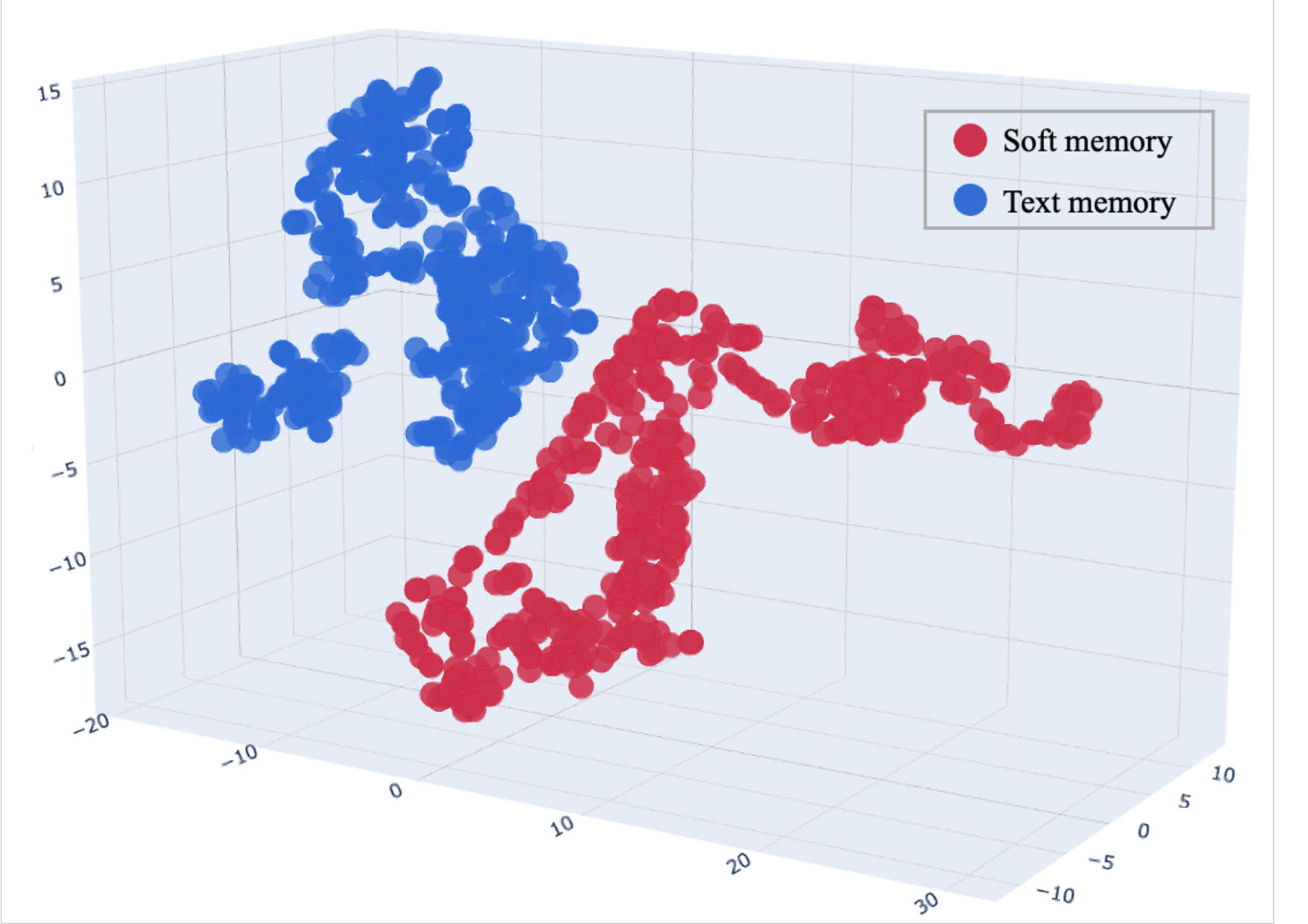}
    \caption{Visualization of soft memory tokens (red) and text memory tokens (blue). The two types occupy non-overlapping regions, confirming that the orthogonality constraint keeps the two channels distinct.}
    \label{fig:tsne}
\vspace{-20pt}
\end{wrapfigure}

To further verify that the latent channel encodes information genuinely distinct from the text channel, we apply t-SNE to visualize the hidden representations of soft and text memory tokens across episodes (Figure~\ref{fig:tsne}). The two token types occupy completely non-overlapping regions, confirming that the orthogonality constraint effectively prevents the latent channel from collapsing into a copy of the text representation.

\paragraph{Efficiency Analysis.}
As shown in Table~\ref{tab:efficiency}, We evaluate AMMI and SCMC on the AlfWorld benchmark, with all experiments conducted on an A100 GPU using vLLM for efficient inference. The results show that SCMC significantly outperforms AMMI, improving the success rate from 35.03\% to 82.16\% while reducing the executor’s computational burden. In particular, SCMC reduces executor input tokens by over 50\% (2481.1 $\rightarrow$ 1003.2) and lowers executor latency from 0.30s to 0.12s, indicating more efficient and focused reasoning. Although SCMC introduces a lightweight memory compilation step, this overhead is offset by the reduced execution cost, resulting in lower overall latency.
\begin{table}[h]\fontsize{8}{10}\selectfont
\centering
% \vspace{-5mm}
\captionsetup{justification=centering}
\caption{Efficiency comparison on AlfWorld.}
\label{tab:efficiency}
\begin{tabular}{lcccccc}
\toprule
& \multicolumn{2}{c}{Latency (s)} 
& \multicolumn{3}{c}{Tokens} 
& Success $\uparrow$ \\
\cmidrule(lr){2-3} \cmidrule(lr){4-6}
Method 
& Compiler & Executor & Exec In & Exec Out & Compiler Total & Rate (\%) \\
\midrule
AMMI & -      & 0.30  & 2481.1 & 18.2  & -  & 35.03 \\
SCMC & 0.093  & \textbf{0.12}  & \textbf{1003.2} & \textbf{8.2} & 2081.4 & \textbf{82.16} \\
\bottomrule
\end{tabular}
\end{table}

% These results suggest that selectively compiling memory into state-relevant context is more effective than directly injecting the full memory, yielding both better performance and efficiency.

\vspace{-5mm}
\section{Conclusion}
We present \textbf{MemCompiler}, a state-conditioned memory  paradigm that replaces the prevailing Ahead-of-time Monolithic Memory Injection with step-wise compilation. A learned Memory Compiler grounds compilation in a structured Brief State, while Soft-Mem delivers perceptual content beyond text. Across a wide range of embodied benchmarks, MemCompiler consistently improves every open-source executor, narrows the gap to frontier closed-source models, and reduces executor cost. These results suggest that progress in memory for embodied agents will increasingly depend on \textit{when} and \textit{how} memory is delivered, not solely on \textit{what} is stored.

{
\small
\bibliographystyle{unsrt}
\bibliography{neurips_2026}

\begin{thebibliography}{10}

\bibitem{agrawal2023physical}
Ayush Agrawal, Raghav Prabhakar, Anirudh Goyal, and Dianbo Liu.
\newblock Physical reasoning and object planning for household embodied agents.
\newblock {\em arXiv preprint arXiv:2311.13577}, 2023.

\bibitem{feng2026multi}
Zhaohan Feng, Ruiqi Xue, Lei Yuan, Yang Yu, Ning Ding, Meiqin Liu, Bingzhao Gao, Jian Sun, Xinhu Zheng, and Gang Wang.
\newblock Multi-agent embodied ai: Advances and future directions.
\newblock {\em Science China Information Sciences}, 69(5):151202, 2026.

\bibitem{team2025gemini}
Gemini~Robotics Team, Abbas Abdolmaleki, Saminda Abeyruwan, Joshua Ainslie, Jean-Baptiste Alayrac, Montserrat~Gonzalez Arenas, Ashwin Balakrishna, Nathan Batchelor, Alex Bewley, Jeff Bingham, et~al.
\newblock Gemini robotics 1.5: Pushing the frontier of generalist robots with advanced embodied reasoning, thinking, and motion transfer.
\newblock {\em arXiv preprint arXiv:2510.03342}, 2025.

\bibitem{glocker2025llm}
Marc Glocker, Peter H{\"o}nig, Matthias Hirschmanner, and Markus Vincze.
\newblock Llm-empowered embodied agent for memory-augmented task planning in household robotics.
\newblock {\em arXiv preprint arXiv:2504.21716}, 2025.

\bibitem{szot2021habitat}
Andrew Szot, Alexander Clegg, Eric Undersander, Erik Wijmans, Yili Zhao, John Turner, Noah Maestre, Mustafa Mukadam, Devendra~Singh Chaplot, Oleksandr Maksymets, et~al.
\newblock Habitat 2.0: Training home assistants to rearrange their habitat.
\newblock {\em Advances in neural information processing systems}, 34:251--266, 2021.

\bibitem{szot2023large}
Andrew Szot, Max Schwarzer, Harsh Agrawal, Bogdan Mazoure, Rin Metcalf, Walter Talbott, Natalie Mackraz, R~Devon Hjelm, and Alexander~T Toshev.
\newblock Large language models as generalizable policies for embodied tasks.
\newblock In {\em The Twelfth International Conference on Learning Representations}, 2023.

\bibitem{zheng2022vlmbench}
Kaizhi Zheng, Xiaotong Chen, Odest~Chadwicke Jenkins, and Xin Wang.
\newblock Vlmbench: A compositional benchmark for vision-and-language manipulation.
\newblock {\em Advances in Neural Information Processing Systems}, 35:665--678, 2022.

\bibitem{cao2025remember}
Zouying Cao, Jiaji Deng, Li~Yu, Weikang Zhou, Zhaoyang Liu, Bolin Ding, and Hai Zhao.
\newblock Remember me, refine me: A dynamic procedural memory framework for experience-driven agent evolution.
\newblock {\em arXiv preprint arXiv:2512.10696}, 2025.

\bibitem{yan2025memory}
Sikuan Yan, Xiufeng Yang, Zuchao Huang, Ercong Nie, Zifeng Ding, Zonggen Li, Xiaowen Ma, Jinhe Bi, Kristian Kersting, Jeff~Z Pan, et~al.
\newblock Memory-r1: Enhancing large language model agents to manage and utilize memories via reinforcement learning.
\newblock {\em arXiv preprint arXiv:2508.19828}, 2025.

\bibitem{ouyang2025reasoningbank}
Siru Ouyang, Jun Yan, I~Hsu, Yanfei Chen, Ke~Jiang, Zifeng Wang, Rujun Han, Long~T Le, Samira Daruki, Xiangru Tang, et~al.
\newblock Reasoningbank: Scaling agent self-evolving with reasoning memory.
\newblock {\em arXiv preprint arXiv:2509.25140}, 2025.

\bibitem{rezazadeh2024isolated}
Alireza Rezazadeh, Zichao Li, Wei Wei, and Yujia Bao.
\newblock From isolated conversations to hierarchical schemas: Dynamic tree memory representation for llms.
\newblock {\em arXiv preprint arXiv:2410.14052}, 2024.

\bibitem{limbacher2020h}
Thomas Limbacher and Robert Legenstein.
\newblock H-mem: Harnessing synaptic plasticity with hebbian memory networks.
\newblock {\em Advances in Neural Information Processing Systems}, 33:21627--21637, 2020.

\bibitem{chhikara2025mem0}
Prateek Chhikara, Dev Khant, Saket Aryan, Taranjeet Singh, and Deshraj Yadav.
\newblock Mem0: Building production-ready ai agents with scalable long-term memory.
\newblock {\em arXiv preprint arXiv:2504.19413}, 2025.

\bibitem{wu2025sgmem}
Yaxiong Wu, Yongyue Zhang, Sheng Liang, and Yong Liu.
\newblock Sgmem: Sentence graph memory for long-term conversational agents.
\newblock {\em arXiv preprint arXiv:2509.21212}, 2025.

\bibitem{hu2025hiagent}
Mengkang Hu, Tianxing Chen, Qiguang Chen, Yao Mu, Wenqi Shao, and Ping Luo.
\newblock Hiagent: Hierarchical working memory management for solving long-horizon agent tasks with large language model.
\newblock In {\em Proceedings of the 63rd Annual Meeting of the Association for Computational Linguistics (Volume 1: Long Papers)}, pages 32779--32798, 2025.

\bibitem{long2025seeing}
Lin Long, Yichen He, Wentao Ye, Yiyuan Pan, Yuan Lin, Hang Li, Junbo Zhao, and Wei Li.
\newblock Seeing, listening, remembering, and reasoning: A multimodal agent with long-term memory.
\newblock {\em arXiv preprint arXiv:2508.09736}, 2025.

\bibitem{openclaw2026}
{OpenClaw}.
\newblock Openclaw: The ai that actually does things.
\newblock \url{https://openclaw.ai/}, 2026.
\newblock Accessed: 2026-05-04.

\bibitem{anthropic_claude_code}
{Anthropic}.
\newblock Claude code: Ai-powered coding assistant for developers.
\newblock \url{https://claude.com/product/claude-code}, 2026.
\newblock Accessed: 2026-05-04.

\bibitem{shridhar2020alfred}
Mohit Shridhar, Jesse Thomason, Daniel Gordon, Yonatan Bisk, Winson Han, Roozbeh Mottaghi, Luke Zettlemoyer, and Dieter Fox.
\newblock Alfred: A benchmark for interpreting grounded instructions for everyday tasks.
\newblock In {\em Proceedings of the IEEE/CVF conference on computer vision and pattern recognition}, pages 10740--10749, 2020.

\bibitem{li2023behavior}
Chengshu Li, Ruohan Zhang, Josiah Wong, Cem Gokmen, Sanjana Srivastava, Roberto Mart{\'\i}n-Mart{\'\i}n, Chen Wang, Gabrael Levine, Michael Lingelbach, Jiankai Sun, et~al.
\newblock Behavior-1k: A benchmark for embodied ai with 1,000 everyday activities and realistic simulation.
\newblock In {\em Conference on Robot Learning}, pages 80--93. PMLR, 2023.

\bibitem{kim2025robot}
Dongyoung Kim, Sumin Park, Huiwon Jang, Jinwoo Shin, Jaehyung Kim, and Younggyo Seo.
\newblock Robot-r1: Reinforcement learning for enhanced embodied reasoning in robotics.
\newblock {\em arXiv preprint arXiv:2506.00070}, 2025.

\bibitem{nasiriany2026robocasa365}
Soroush Nasiriany, Sepehr Nasiriany, Abhiram Maddukuri, and Yuke Zhu.
\newblock Robocasa365: A large-scale simulation framework for training and benchmarking generalist robots.
\newblock {\em arXiv preprint arXiv:2603.04356}, 2026.

\bibitem{srivastava2022behavior}
Sanjana Srivastava, Chengshu Li, Michael Lingelbach, Roberto Mart{\'\i}n-Mart{\'\i}n, Fei Xia, Kent~Elliott Vainio, Zheng Lian, Cem Gokmen, Shyamal Buch, Karen Liu, et~al.
\newblock Behavior: Benchmark for everyday household activities in virtual, interactive, and ecological environments.
\newblock In {\em Conference on robot learning}, pages 477--490. PMLR, 2022.

\bibitem{shao2024deepseekmath}
Zhihong Shao, Peiyi Wang, Qihao Zhu, Runxin Xu, Junxiao Song, Xiao Bi, Haowei Zhang, Mingchuan Zhang, YK~Li, Yang Wu, et~al.
\newblock Deepseekmath: Pushing the limits of mathematical reasoning in open language models.
\newblock {\em arXiv preprint arXiv:2402.03300}, 2024.

\bibitem{park2023generative}
Joon~Sung Park, Joseph O'Brien, Carrie~Jun Cai, Meredith~Ringel Morris, Percy Liang, and Michael~S Bernstein.
\newblock Generative agents: Interactive simulacra of human behavior.
\newblock In {\em Proceedings of the 36th annual acm symposium on user interface software and technology}, pages 1--22, 2023.

\bibitem{wang2024mobile}
Junyang Wang, Haiyang Xu, Haitao Jia, Xi~Zhang, Ming Yan, Weizhou Shen, Ji~Zhang, Fei Huang, and Jitao Sang.
\newblock Mobile-agent-v2: Mobile device operation assistant with effective navigation via multi-agent collaboration.
\newblock {\em Advances in Neural Information Processing Systems}, 37:2686--2710, 2024.

\bibitem{shinn2023reflexion}
Noah Shinn, Federico Cassano, Ashwin Gopinath, Karthik Narasimhan, and Shunyu Yao.
\newblock Reflexion: Language agents with verbal reinforcement learning.
\newblock {\em Advances in neural information processing systems}, 36:8634--8652, 2023.

\bibitem{packer2023memgpt}
Charles Packer, Vivian Fang, Shishir\_G Patil, Kevin Lin, Sarah Wooders, and Joseph\_E Gonzalez.
\newblock Memgpt: towards llms as operating systems.
\newblock 2023.

\bibitem{zhao2024expel}
Andrew Zhao, Daniel Huang, Quentin Xu, Matthieu Lin, Yong-Jin Liu, and Gao Huang.
\newblock Expel: Llm agents are experiential learners.
\newblock In {\em Proceedings of the AAAI Conference on Artificial Intelligence}, volume~38, pages 19632--19642, 2024.

\bibitem{gutierrez2024hipporag}
Bernal~J Guti{\'e}rrez, Yiheng Shu, Yu~Gu, Michihiro Yasunaga, and Yu~Su.
\newblock Hipporag: Neurobiologically inspired long-term memory for large language models.
\newblock {\em Advances in neural information processing systems}, 37:59532--59569, 2024.

\bibitem{kadavy2021digital}
David Kadavy.
\newblock {\em Digital Zettelkasten: Principles, Methods, \& Examples}.
\newblock Kadavy, Inc., 2021.

\bibitem{ahrens2022take}
S{\"o}nke Ahrens.
\newblock {\em How to take smart notes: One simple technique to boost writing, learning and thinking}.
\newblock S{\"o}nke Ahrens, 2022.

\bibitem{xu2025mem}
Wujiang Xu, Zujie Liang, Kai Mei, Hang Gao, Juntao Tan, and Yongfeng Zhang.
\newblock A-mem: Agentic memory for llm agents.
\newblock {\em arXiv preprint arXiv:2502.12110}, 2025.

\bibitem{anthropic2026claudeopus46}
{Anthropic}.
\newblock Introducing claude opus 4.6.
\newblock \url{https://www.anthropic.com/news/claude-opus-4-6}, 2026.
\newblock Online; accessed 2026-05-05.

\bibitem{brown2020language}
Tom Brown, Benjamin Mann, Nick Ryder, Melanie Subbiah, Jared~D Kaplan, Prafulla Dhariwal, Arvind Neelakantan, Pranav Shyam, Girish Sastry, Amanda Askell, et~al.
\newblock Language models are few-shot learners.
\newblock {\em Advances in neural information processing systems}, 33:1877--1901, 2020.

\bibitem{team2023gemini}
Gemini Team, Rohan Anil, Sebastian Borgeaud, Jean-Baptiste Alayrac, Jiahui Yu, Radu Soricut, Johan Schalkwyk, Andrew~M Dai, Anja Hauth, Katie Millican, et~al.
\newblock Gemini: a family of highly capable multimodal models.
\newblock {\em arXiv preprint arXiv:2312.11805}, 2023.

\bibitem{wang2023voyager}
Guanzhi Wang, Yuqi Xie, Yunfan Jiang, Ajay Mandlekar, Chaowei Xiao, Yuke Zhu, Linxi Fan, and Anima Anandkumar.
\newblock Voyager: An open-ended embodied agent with large language models.
\newblock {\em arXiv preprint arXiv:2305.16291}, 2023.

\bibitem{wang2024jarvis}
Zihao Wang, Shaofei Cai, Anji Liu, Yonggang Jin, Jinbing Hou, Bowei Zhang, Haowei Lin, Zhaofeng He, Zilong Zheng, Yaodong Yang, et~al.
\newblock Jarvis-1: Open-world multi-task agents with memory-augmented multimodal language models.
\newblock {\em IEEE Transactions on Pattern Analysis and Machine Intelligence}, 47(3):1894--1907, 2024.

\bibitem{zhang2025g}
Guibin Zhang, Muxin Fu, Guancheng Wan, Miao Yu, Kun Wang, and Shuicheng Yan.
\newblock G-memory: Tracing hierarchical memory for multi-agent systems.
\newblock {\em arXiv preprint arXiv:2506.07398}, 2025.

\bibitem{wei2025evo}
Tianxin Wei, Noveen Sachdeva, Benjamin Coleman, Zhankui He, Yuanchen Bei, Xuying Ning, Mengting Ai, Yunzhe Li, Jingrui He, Ed~H Chi, et~al.
\newblock Evo-memory: Benchmarking llm agent test-time learning with self-evolving memory.
\newblock {\em arXiv preprint arXiv:2511.20857}, 2025.

\bibitem{yan2025general}
BY~Yan, Chaofan Li, Hongjin Qian, Shuqi Lu, and Zheng Liu.
\newblock General agentic memory via deep research.
\newblock {\em arXiv preprint arXiv:2511.18423}, 2025.

\bibitem{lei2025robomemory}
Mingcong Lei, Honghao Cai, Zezhou Cui, Liangchen Tan, Junkun Hong, Gehan Hu, Shuangyu Zhu, Yimou Wu, Shaohan Jiang, Ge~Wang, et~al.
\newblock Robomemory: A brain-inspired multi-memory agentic framework for lifelong learning in physical embodied systems.
\newblock In {\em NeurIPS 2025 Workshop on Space in Vision, Language, and Embodied AI}, 2025.

\bibitem{zhang2026nextmem}
Zeyu Zhang, Rui Li, Xiaoyan Zhao, Yang Zhang, Wenjie Wang, Xu~Chen, and Tat-Seng Chua.
\newblock Nextmem: Towards latent factual memory for llm-based agents.
\newblock {\em arXiv preprint arXiv:2603.15634}, 2026.

\bibitem{zhang2025memgen}
Guibin Zhang, Muxin Fu, and Shuicheng Yan.
\newblock Memgen: Weaving generative latent memory for self-evolving agents.
\newblock {\em arXiv preprint arXiv:2509.24704}, 2025.

\bibitem{yu2025vismem}
Xinlei Yu, Chengming Xu, Guibin Zhang, Zhangquan Chen, Yudong Zhang, Yongbo He, Peng-Tao Jiang, Jiangning Zhang, Xiaobin Hu, and Shuicheng Yan.
\newblock Vismem: Latent vision memory unlocks potential of vision-language models.
\newblock {\em arXiv preprint arXiv:2511.11007}, 2025.

\bibitem{shridhar2020alfworld}
Mohit Shridhar, Xingdi Yuan, Marc-Alexandre C{\^o}t{\'e}, Yonatan Bisk, Adam Trischler, and Matthew Hausknecht.
\newblock Alfworld: Aligning text and embodied environments for interactive learning.
\newblock {\em arXiv preprint arXiv:2010.03768}, 2020.

\bibitem{yang2025embodiedbench}
Rui Yang, Hanyang Chen, Junyu Zhang, Mark Zhao, Cheng Qian, Kangrui Wang, Qineng Wang, Teja~Venkat Koripella, Marziyeh Movahedi, Manling Li, et~al.
\newblock Embodiedbench: Comprehensive benchmarking multi-modal large language models for vision-driven embodied agents.
\newblock {\em arXiv preprint arXiv:2502.09560}, 2025.

\bibitem{wang2022scienceworld}
Ruoyao Wang, Peter Jansen, Marc-Alexandre C{\^o}t{\'e}, and Prithviraj Ammanabrolu.
\newblock Scienceworld: Is your agent smarter than a 5th grader?
\newblock In {\em Proceedings of the 2022 Conference on Empirical Methods in Natural Language Processing}, pages 11279--11298, 2022.

\bibitem{langchain2026langmem}
{LangChain AI}.
\newblock Langmem: A framework for long-term memory in llm agents.
\newblock \url{https://github.com/langchain-ai/langmem}, 2026.
\newblock Accessed: 2026-05-04.

\bibitem{bai2025qwen25vltechnicalreport}
Shuai Bai, Keqin Chen, Xuejing Liu, Jialin Wang, Wenbin Ge, Sibo Song, Kai Dang, Peng Wang, Shijie Wang, Jun Tang, Humen Zhong, Yuanzhi Zhu, Mingkun Yang, Zhaohai Li, Jianqiang Wan, Pengfei Wang, Wei Ding, Zheren Fu, Yiheng Xu, Jiabo Ye, Xi~Zhang, Tianbao Xie, Zesen Cheng, Hang Zhang, Zhibo Yang, Haiyang Xu, and Junyang Lin.
\newblock Qwen2.5-vl technical report, 2025.

\bibitem{bai2025qwen3}
Shuai Bai, Yuxuan Cai, Ruizhe Chen, Keqin Chen, Xionghui Chen, Zesen Cheng, Lianghao Deng, Wei Ding, Chang Gao, Chunjiang Ge, et~al.
\newblock Qwen3-vl technical report.
\newblock {\em arXiv preprint arXiv:2511.21631}, 2025.

\bibitem{cote2018textworld}
Marc-Alexandre C{\^o}t{\'e}, Akos K{\'a}d{\'a}r, Xingdi Yuan, Ben Kybartas, Tavian Barnes, Emery Fine, James Moore, Matthew Hausknecht, Layla El~Asri, Mahmoud Adada, et~al.
\newblock Textworld: A learning environment for text-based games.
\newblock In {\em Workshop on Computer Games}, pages 41--75. Springer, 2018.

\bibitem{kolve2017ai2}
Eric Kolve, Roozbeh Mottaghi, Winson Han, Eli VanderBilt, Luca Weihs, Alvaro Herrasti, Matt Deitke, Kiana Ehsani, Daniel Gordon, Yuke Zhu, et~al.
\newblock Ai2-thor: An interactive 3d environment for visual ai.
\newblock {\em arXiv preprint arXiv:1712.05474}, 2017.

\bibitem{calli2015ycb}
Berk Calli, Arjun Singh, Aaron Walsman, Siddhartha Srinivasa, Pieter Abbeel, and Aaron~M Dollar.
\newblock The ycb object and model set: Towards common benchmarks for manipulation research.
\newblock In {\em 2015 international conference on advanced robotics (ICAR)}, pages 510--517. IEEE, 2015.

\bibitem{shang2024agentsquare}
Yu~Shang, Yu~Li, Keyu Zhao, Likai Ma, Jiahe Liu, Fengli Xu, and Yong Li.
\newblock Agentsquare: Automatic llm agent search in modular design space.
\newblock In {\em The Thirteenth International Conference on Learning Representations}, 2024.

\bibitem{hu2022lora}
Edward~J Hu, Yelong Shen, Phillip Wallis, Zeyuan Allen-Zhu, Yuanzhi Li, Shean Wang, Liang Wang, Weizhu Chen, et~al.
\newblock Lora: Low-rank adaptation of large language models.
\newblock {\em Iclr}, 1(2):3, 2022.

\end{thebibliography}
}

%%%%%%%%%%%%%%%%%%%%%%%%%%%%%%%%%%%%%%%%%%%%%%%%%%%%%%%%%%%%

\appendix

\section{Progress Rate on EmbodiedBench}
\label{app:embodiedbench-progress}
Table~\ref{app:table1} reports \textit{Progress Rate}, which credits partial subgoal completion, as a complement to the strict Success Rate in Table~\ref{table1}. The Progress view confirms the trends observed on Success Rate while making the gap between methods more visible. On open-source backbones, MemCompiler is again the only method that improves over No Mem on every (executor, dataset) cell, with relative gains of +54.3\% on Qwen-2.5-VL-32B / EB-ALFRED and +63.8\% on Qwen-3-VL-32B / EB-ALFRED. The AMMI-style baselines that collapse on Success Rate also collapse here (e.g., G-Mem $-53.1\%$, A-Mem $-61.8\%$ on Qwen-2.5-VL-32B / EB-ALFRED), and MemGen continues to underperform across both Qwen-VL backbones, reinforcing that latent representations alone, without state-conditioned compilation, do not transfer to embodied execution. On closed-source backbones, AMMI baselines yield smaller relative gains under Progress Rate than under Success Rate (e.g., G-Mem on GPT-5 / EB-ALFRED rises by +37.5\% rather than +42.1\%), consistent with Progress already crediting the partial completions that No Mem can sometimes reach. Overall, Progress Rate offers a finer-grained but qualitatively identical picture: state-conditioned compilation helps every executor, while static injection helps only when the executor is large enough to filter it.

\begin{table}[h]\fontsize{7.5}{10}\selectfont %字号，行间距
\caption{Progress Rate (\%) on EmbodiedBench (EB-ALFRED and EB-Habitat) (\%). Capability dimensions: Base (base ability), Com.\ (commonsense reasoning), Comp.\ (complex instruction following), Vis.\ (visual recognition), Spat.\ (spatial reasoning), Long (long-horizon planning).}
% \vspace{-3mm}
\label{app:table1}
\renewcommand{\tabcolsep}{1pt} % 将列间距设为 3pt
\begin{tabular}{cccccccccccccccc}
\toprule
\multirow{2}{*}{Model}                                                      & \multicolumn{1}{c|}{\multirow{2}{*}{Method}} & \multicolumn{7}{c|}{EB-ALFRED}                                                                                                                                      & \multicolumn{7}{c}{EB-Habitat}                                                                                                                                 \\ \cline{3-16} 
& \multicolumn{1}{c|}{}                        & Avg.                  & Base                 & Com.               & Comp.              & Vis.               & Spat.              & \multicolumn{1}{c|}{Long} & Avg.                  & Base                 & Com.               & Comp.              & Vis.               & Spat.              & Long                 \\ \hline
\multicolumn{16}{c}{Close-Source MLLMs}      \\ \hline
\multirow{5}{*}{GPT-5.2}                                                      & \multicolumn{1}{c|}{No Mem}                  &    41.29  {\scriptsize \base{0}}                &66.99            &48.33             &    42.33                  &      43.92                &     36.67                 & \multicolumn{1}{c|}{9.5}     &     52.85   {\scriptsize \base{0}}                 &     81.75                 &          44.5            &   52                   &        46.83              &     59.83                 &         32.20             \\
& \multicolumn{1}{c|}{Mem0}                    &        53.24  {\scriptsize \up{28.9}}            &         57.67             &        65.33              &     67.33                 &     56.12                 &       47.99               & \multicolumn{1}{c|}{25}     &        53.53  {\scriptsize \up{1.3}}            &        81              &        45              &       48               &        58              &       62.25               &          26.95            \\
& \multicolumn{1}{c|}{G-Mem}                   &    56.78   {\scriptsize \up{37.5}}               &          57.17            &      63.67                &       65.33               &       55.5               &         52             & \multicolumn{1}{c|}{47}     &       55.2  {\scriptsize \up{4.4}}             &84.5            &          46            &           43           &         63.67             &        59.08              &        34.95              \\
 & \multicolumn{1}{c|}{A-Mem}                    &        57.8 {\scriptsize \up{40.0}}             &             58.33         &      58.83                &         59.33             &       48.83               &       57               & \multicolumn{1}{c|}{64.5}     &       44.77  {\scriptsize \down{15.3}}             &        70              &         33             &     35                 &          56.67            &         55.83             &          18.12            \\
  & \multicolumn{1}{c|}{LangMem}                    &        57.94    {\scriptsize \up{40.3}}          &         79.83             &    63                  &            66.33          &        55.17              &      51.33                & \multicolumn{1}{c|}{32}     &           55.06    {\scriptsize \up{4.2}}       &         86             &       49               &       48.51               &           63.17           &      61.67                &   22.03                   \\
                                                                            \hline
\multirow{5}{*}{\begin{tabular}[c]{@{}c@{}}Gemini-3\\ -flash\end{tabular}}                                                     & \multicolumn{1}{c|}{No Mem}                  &   68.47   {\scriptsize \base{0}}                   &           76.67           &        65.33              &    67.33                  &        72.5              &     67                 & \multicolumn{1}{c|}{62}     & 64.2    {\scriptsize \base{0}}                    &        88.5              &      61                &           61           &      69.92                &            59.92          &       44.86               \\
& \multicolumn{1}{c|}{Mem0}                    & \multicolumn{1}{c}{83.33 {\scriptsize \up{21.7}}} & \multicolumn{1}{c}{85} & \multicolumn{1}{c}{84.66} & \multicolumn{1}{c}{86.33} & \multicolumn{1}{c}{84.83} & \multicolumn{1}{c}{80.67} & \multicolumn{1}{l|}{78.5}     & \multicolumn{1}{c}{70.22 {\scriptsize \up{9.4}}} & \multicolumn{1}{c}{89.25} & \multicolumn{1}{c}{ 75} & \multicolumn{1}{c}{61} & \multicolumn{1}{c}{79.08} & \multicolumn{1}{c}{61.58} & \multicolumn{1}{c}{55.42} \\
& \multicolumn{1}{c|}{G-Mem}                   &      84.23  {\scriptsize \up{23.0}}              &           85.71           &    85.33                  &      83.67                &     84.5                 &          82.67            & \multicolumn{1}{c|}{83.5}     &       73.04    {\scriptsize \up{13.8}}           &      93.75                &        73              &        62.36              &       80.65               &            80.19          &        48.32              \\ 
 & \multicolumn{1}{c|}{A-Mem}                    &      86.69   {\scriptsize \up{26.6}}             &         92             &      82.83                &        89              &      88.17                &           86.67           & \multicolumn{1}{c|}{81.5}     &       75.5  {\scriptsize \up{17.6}}             &    98.5                  &    75.5                  & 73.17                     &         83.17             &             82.11         &        40.6              \\
  & \multicolumn{1}{c|}{LangMem}                    &         83.01   {\scriptsize \up{21.2}}          &       85.67               &         79.22             &          80.56            &              86.14        &              85.5        & \multicolumn{1}{c|}{81}     &            71.49   {\scriptsize \up{11.4}}       &            89          &          74            &           66           &        84.23              &          83            &        32.53              \\\hline
\multicolumn{16}{c}{Open-Source MLLMs}                                                                                                                                      \\ \hline
\multirow{7}{*}{\begin{tabular}[c]{@{}c@{}}Qwen-2.5\\ -VL-32B\end{tabular}} & \multicolumn{1}{c|}{No Mem}      &             28.86    {\scriptsize \base{0}}        &          33.67            &          32.33            &         36.67             &   25.83                   &       31.67               & \multicolumn{1}{c|}{13}     &        48.22    {\scriptsize \base{0}}             &      80.75                &   33                   &      60                &     34.58                 &              56.83        &        24.19              \\
& \multicolumn{1}{c|}{G-Mem}                   & \multicolumn{1}{c}{13.53 {\scriptsize \down{53.1}}} & \multicolumn{1}{c}{10} & \multicolumn{1}{c}{20.67} & \multicolumn{1}{c}{14.67} & \multicolumn{1}{c}{21.67} & \multicolumn{1}{c}{11.67} & \multicolumn{1}{l|}{2.5}     & \multicolumn{1}{c}{25.31 {\scriptsize \down{47.5}}} & \multicolumn{1}{c}{50.33} & \multicolumn{1}{c}{19} & \multicolumn{1}{c}{21.5} & \multicolumn{1}{c}{16.17} & \multicolumn{1}{c}{40.17} & \multicolumn{1}{c}{4.67} \\
& \multicolumn{1}{c|}{Mem0}                    & \multicolumn{1}{c}{23.44 {\scriptsize \down{18.8}}} & \multicolumn{1}{c}{28} & \multicolumn{1}{c}{25.83} & \multicolumn{1}{c}{37.33} & \multicolumn{1}{c}{25} & \multicolumn{1}{c}{17} & \multicolumn{1}{l|}{7.5}     & \multicolumn{1}{c}{51.97 {\scriptsize \up{7.8}}} & \multicolumn{1}{c}{74.33} & \multicolumn{1}{c}{42.5} & \multicolumn{1}{c}{58.86} & \multicolumn{1}{c}{45.25} & \multicolumn{1}{c}{57.5} & \multicolumn{1}{c}{33.36} \\
& \multicolumn{1}{c|}{A-Mem}                   & \multicolumn{1}{c}{11.02 {\scriptsize \down{61.8}}} & \multicolumn{1}{c}{12.33} & \multicolumn{1}{c}{15.67} & \multicolumn{1}{c}{15.33} & \multicolumn{1}{c}{15.99} & \multicolumn{1}{c}{6.3} & \multicolumn{1}{l|}{0.5}     & \multicolumn{1}{c}{17.56 {\scriptsize \down{63.6}}} & \multicolumn{1}{c}{37.25} & \multicolumn{1}{c}{15.5} & \multicolumn{1}{c}{10} & \multicolumn{1}{c}{15.5} & \multicolumn{1}{c}{21.75} & \multicolumn{1}{c}{5.33} \\
& \multicolumn{1}{c|}{Langmem}                    & \multicolumn{1}{c}{22.69 {\scriptsize \down{21.4}}} & \multicolumn{1}{c}{29.17} & \multicolumn{1}{c}{27.33} & \multicolumn{1}{c}{35.49} & \multicolumn{1}{c}{20} & \multicolumn{1}{c}{19.67} & \multicolumn{1}{l|}{4.5}     & \multicolumn{1}{c}{51.31 {\scriptsize \up{6.4}}} & \multicolumn{1}{c}{83.25} & \multicolumn{1}{c}{39} & \multicolumn{1}{c}{52} & \multicolumn{1}{c}{46} & \multicolumn{1}{c}{55.5} & \multicolumn{1}{c}{32.14} \\
& \multicolumn{1}{c|}{MemGen}                    & \multicolumn{1}{c}{15.4 {\scriptsize \down{46.6}}} & \multicolumn{1}{c}{19.64} & \multicolumn{1}{c}{17.31} & \multicolumn{1}{c}{24.22} & \multicolumn{1}{c}{12.67} & \multicolumn{1}{c}{15.33} & \multicolumn{1}{l|}{3.2}     & \multicolumn{1}{c}{31.43 {\scriptsize \down{34.8}}} & \multicolumn{1}{c}{57.91} & \multicolumn{1}{c}{15} & \multicolumn{1}{c}{25} & \multicolumn{1}{c}{24.33} & \multicolumn{1}{c}{56.17} & \multicolumn{1}{c}{10.17} \\
& \multicolumn{1}{c|}{Ours}                    &           44.53   {\scriptsize \up{54.3}}        &           44.67           &        61.52              &  62.33                    &           45.89           &    42.45                  & \multicolumn{1}{c|}{10.37}     &          63.69  {\scriptsize \up{32.1}}          &        86.75              &       62.5               &          59            &       65.41               &     61.57                 &         46.93             \\ \hline
\multirow{7}{*}{\begin{tabular}[c]{@{}c@{}}Qwen-3\\-VL-32B\end{tabular}}   & \multicolumn{1}{c|}{No Mem}                  &        29.69     {\scriptsize \base{0}}            &        36              &       31.83               &       38.17               &         35.5             &      26.17                & \multicolumn{1}{c|}{10.5}     &          45.89   {\scriptsize \base{0}}            &         82.67             &    35.5                  &           42           &           40.42           &           47.83           &            26.97          \\
& \multicolumn{1}{c|}{G-Mem}                   & \multicolumn{1}{c}{39.08 {\scriptsize \up{31.6}}} & \multicolumn{1}{c}{46} & \multicolumn{1}{c}{44.5} & \multicolumn{1}{c}{46.5} & \multicolumn{1}{c}{38} & \multicolumn{1}{c}{30} & \multicolumn{1}{l|}{29.5}     & \multicolumn{1}{c}{56.27 {\scriptsize \up{22.6}}} & \multicolumn{1}{c}{91.33} & \multicolumn{1}{c}{35.5} & \multicolumn{1}{c}{47.5} & \multicolumn{1}{c}{57.92} & \multicolumn{1}{c}{57.5} & \multicolumn{1}{c}{47.85} \\
& \multicolumn{1}{c|}{Mem0}                    & \multicolumn{1}{c}{41.08 {\scriptsize \up{38.4}}} & \multicolumn{1}{c}{49.17} & \multicolumn{1}{c}{50.83} & \multicolumn{1}{c}{52.33} & \multicolumn{1}{c}{38.33} & \multicolumn{1}{c}{36.33} & \multicolumn{1}{l|}{19.5}     & \multicolumn{1}{c}{50.03 {\scriptsize \up{9.0}}} & \multicolumn{1}{c}{91.17} & \multicolumn{1}{c}{31.5} & \multicolumn{1}{c}{34.5} & \multicolumn{1}{c}{48.08} & \multicolumn{1}{c}{56.67} & \multicolumn{1}{c}{38.28} \\
& \multicolumn{1}{c|}{A-Mem}                   & \multicolumn{1}{c}{42.45 {\scriptsize \up{43.0}}} & \multicolumn{1}{c}{55.17} & \multicolumn{1}{c}{53.83} & \multicolumn{1}{c}{45.33} & \multicolumn{1}{c}{41.67} & \multicolumn{1}{c}{42.67} & \multicolumn{1}{l|}{16}     & \multicolumn{1}{c}{48.59 {\scriptsize \up{5.9}}} & \multicolumn{1}{c}{89.83} & \multicolumn{1}{c}{20.5} & \multicolumn{1}{c}{52.5} & \multicolumn{1}{c}{29.67} & \multicolumn{1}{c}{57.33} & \multicolumn{1}{c}{41.69} \\
& \multicolumn{1}{c|}{Langmem}                    & \multicolumn{1}{c}{41.72 {\scriptsize \up{40.5}}} & \multicolumn{1}{c}{47.83} & \multicolumn{1}{c}{52.83} & \multicolumn{1}{c}{54} & \multicolumn{1}{c}{39.67} & \multicolumn{1}{c}{37} & \multicolumn{1}{l|}{19}     & \multicolumn{1}{c}{50.56 {\scriptsize \up{10.2}}} & \multicolumn{1}{c}{90.85} & \multicolumn{1}{c}{34.5} & \multicolumn{1}{c}{36.5} & \multicolumn{1}{c}{47.41} & \multicolumn{1}{c}{55.17} & \multicolumn{1}{c}{38.95} \\
& \multicolumn{1}{c|}{MemGen}                    & \multicolumn{1}{c}{21.62 {\scriptsize \down{27.2}}} & \multicolumn{1}{c}{33.64} & \multicolumn{1}{c}{29.31} & \multicolumn{1}{c}{20.11} & \multicolumn{1}{c}{12.67} & \multicolumn{1}{c}{25.47} & \multicolumn{1}{l|}{8.5}     & \multicolumn{1}{c}{32.1 {\scriptsize \down{30.0}}} & \multicolumn{1}{c}{60.23} & \multicolumn{1}{c}{19.67} & \multicolumn{1}{c}{30.33} & \multicolumn{1}{c}{40.47} & \multicolumn{1}{c}{26.21} & \multicolumn{1}{c}{15.7} \\
& \multicolumn{1}{c|}{Ours}                    &      48.64    {\scriptsize \up{63.8}}            &            56.22          &       60.9               &  56.32                    &        47.67              &   48                   & \multicolumn{1}{c|}{22.75}     &        65.7   {\scriptsize \up{43.2}}           &   93.25                   &        60.5              &         55.5             &     66.33                 &              65.31        &     54.33                 \\ \bottomrule
\end{tabular}
\end{table}

\section{Qualitative Case Studies}
\label{app:case-studies}
\begin{figure}[h]
\centering
\includegraphics[width=0.85\textwidth]{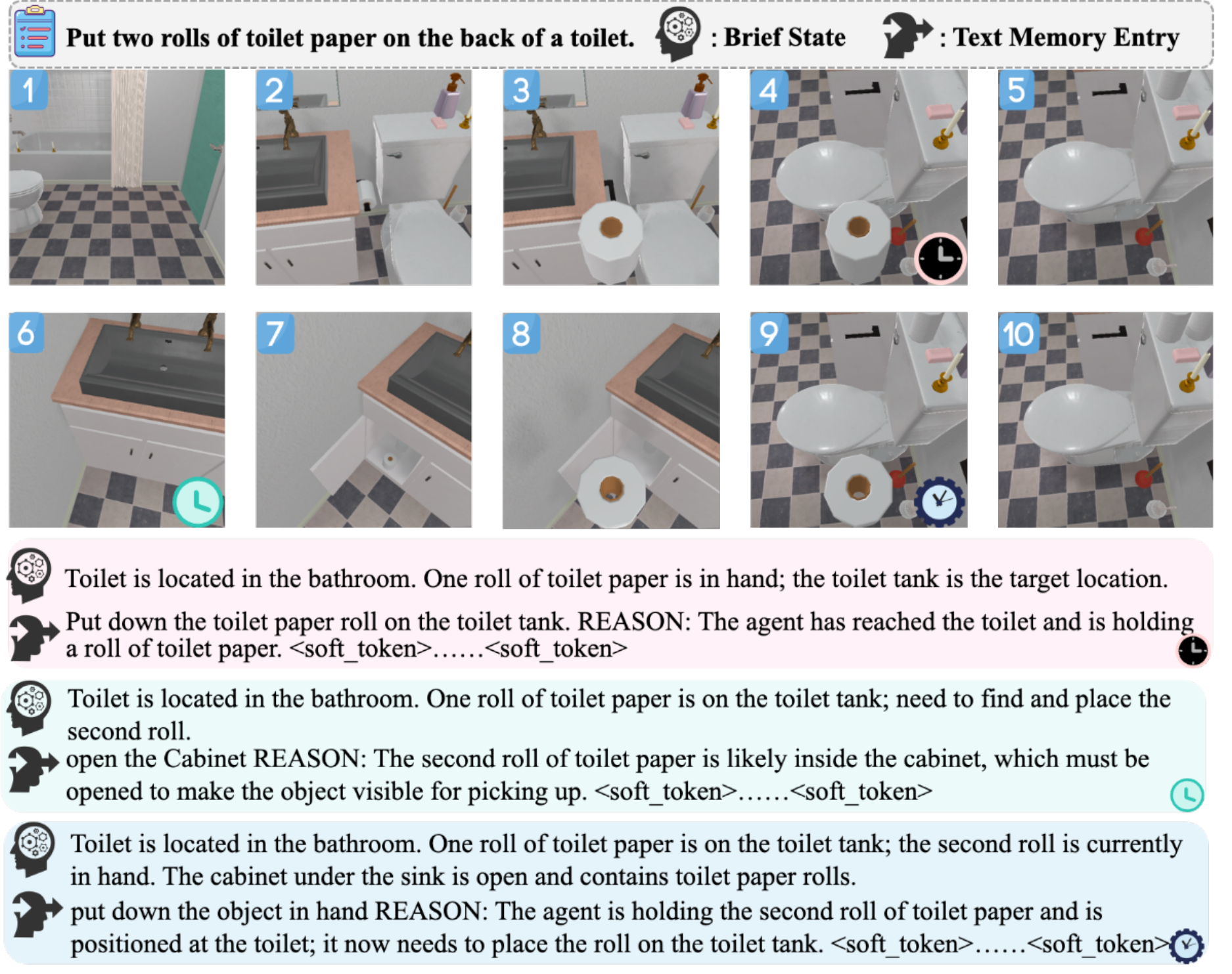}
\caption{Case study on EB-ALFRED. The Memory Compiler advances Brief State and compiled guidance step-by-step as the agent locates, rinses, and places the lettuce.}
\label{fig:case1}
\vspace{-3mm}
\end{figure}

\begin{figure}[h]
\centering
\includegraphics[width=0.8\textwidth]{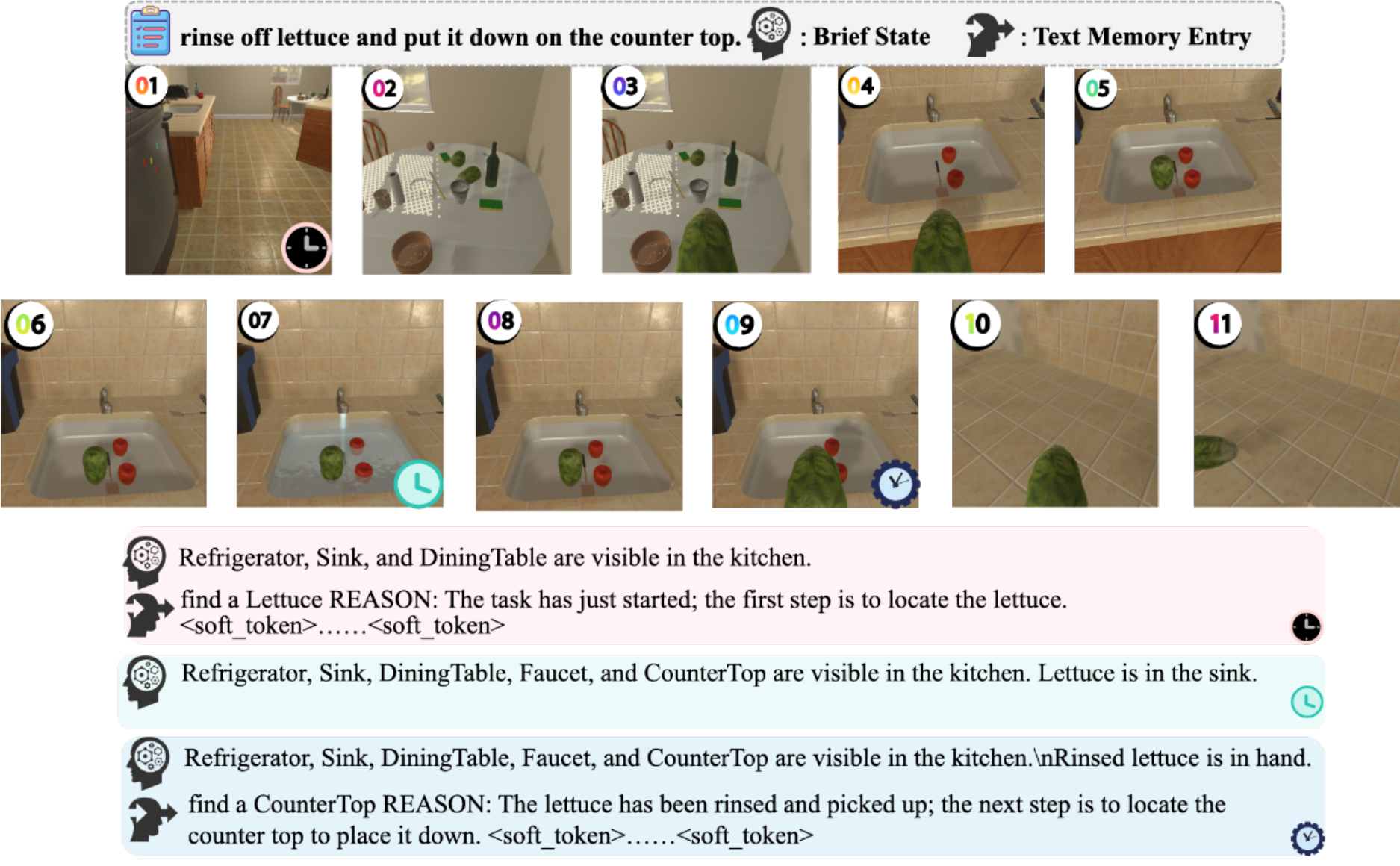}
\caption{Case study on EB-ALFRED. The Memory Compiler detects a stuck-in-loop failure mode in the Brief State and issues a corrective directive that breaks the loop.}
\label{fig:case2}
\vspace{-3mm}
\end{figure}

\begin{figure}[h]
\centering
\includegraphics[width=0.8\textwidth]{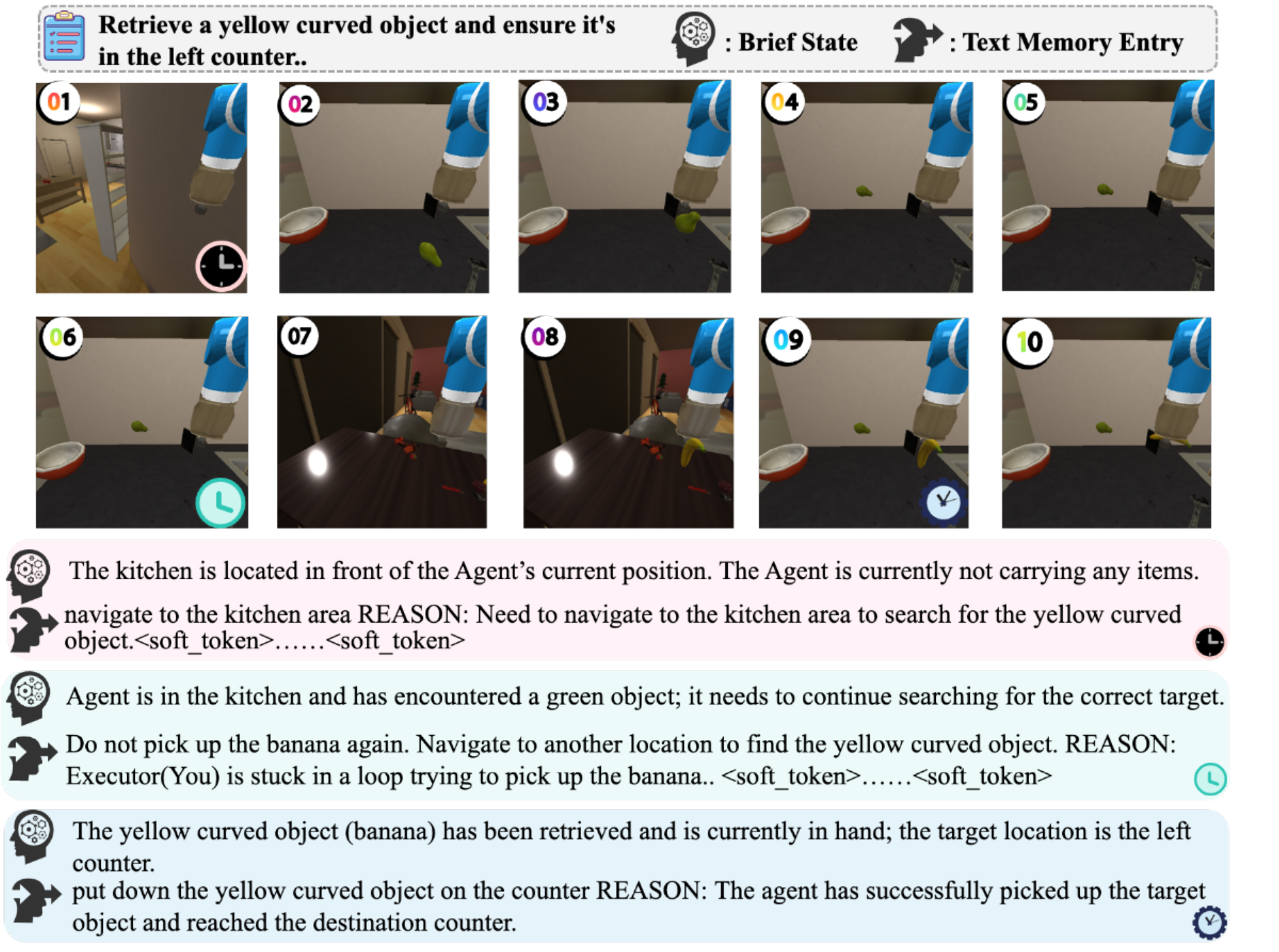}
\caption{Case study on EB-Habitat. The Memory Compiler maintains continuity across multiple subgoals by folding completed steps into the Brief State and selectively withholding memory entries when none is currently applicable.}
\label{fig:case3}
\vspace{-3mm}
\end{figure}

To make MemCompiler's behavior concrete, we visualize three EmbodiedBench episodes (Figures~\ref{fig:case1}--\ref{fig:case3}). Each figure shows the executor's first-person observations across timesteps (top), together with the Brief State  and the text portion of the compiled memory entry emitted by the Memory Compiler; soft tokens (\texttt{<soft\_token>}) are rendered as placeholders for the co-emitted latent channel.

Across the three episodes, the same compilation pattern emerges. The Brief State is not a static summary but a runtime context that the Memory Compiler actively maintains: it records the currently visible scene, folds completed subgoals into accumulated environment beliefs, and registers diagnostic conditions such as \emph{stuck-in-loop} failure modes when the executor's behavior departs from progress. Conditioned on this evolving Brief State, the compiled text entry shifts step-by-step, advancing routine plans (\emph{find a Lettuce} $\to$ \emph{turn off the Faucet} $\to$ \emph{find a CounterTop}), issuing corrective directives when the executor is stuck (\emph{Do not pick up the banana again. Navigate to another location\dots}), and selectively withholding output when no entry in $\mathcal{M}$ is currently applicable. Each compiled entry includes an explicit reason that grounds the action in the Brief State, and is paired with latent soft tokens that carry perceptual context the text channel cannot express. These visualizations confirm the central design intent of MemCompiler: rather than injecting memory once and hoping for relevance, the Memory Compiler delivers state-aligned guidance whose content, timing, and modality follow the agent's evolving execution state.

\section{Experimental Details}
\label{sec:experimental_setup}

\subsection{Benchmarks and Datasets Details}
\label{subsec:datasets_details}

We evaluate our approach across four embodied planning and reasoning benchmarks. To rigorously assess the generalization capabilities of the models, we enforced strict ``zero-leakage'' mechanisms across all benchmarks, ensuring a complete separation between the training and evaluation sets in terms of environment instances, scene layouts, or task instructions.

\paragraph{ALFWorld}
ALFWorld is a text-based interactive environment built upon TextWorld\cite{cote2018textworld} and ALFRED. The agent is required to complete six types of household tasks (e.g., pick and place, examine in light, heat and place) using text commands (e.g., \texttt{go to countertop 1}).
\begin{itemize}
    \item \textbf{Training Set}: Derived from the \texttt{train} split of ALFRED \texttt{json\_2.1.1}, comprising 3,553 task instances across 108 distinct indoor scenes.
    \item \textbf{Evaluation Set}: Utilizes the \texttt{valid\_unseen} split (an out-of-distribution evaluation set), consisting of 134 task instances.
    \item \textbf{Isolation Mechanism}: The evaluation set exclusively evaluates agents in four scenes (FloorPlan 10, 219, 308, 424) that are completely unseen during training. There is strict zero overlap between the training and test sets at both the gamefile and scene levels, rigorously evaluating the model's spatial and interactive generalization in novel environments.
\end{itemize}

\paragraph{EmbodiedBench (EB-ALFRED)}
EB-ALFRED is built upon the AI2-THOR\cite{kolve2017ai2} simulator (v2.1.0) and encompasses seven types of visual household tasks from ALFRED.
\begin{itemize}
    \item \textbf{Evaluation Set}: Employs episodes uniformly distributed across six fine-grained evaluation subsets provided by EmbodiedBench (Base, Common Sense, Complex Instruction, Spatial, Visual Appearance, and Long Horizon). To specifically evaluate diverse capabilities, the instructions in the test set were manually rewritten from the original ALFRED annotations (e.g., substituting specific object names with functional descriptions for the ``Common Sense'' subset, and introducing conditional statements and distractors for ``Complex Instruction'').
    \item \textbf{Training Set}: We used an automated script to scan all available tasks in ALFRED \texttt{json\_2.1.0}, strictly excluding the 213 \texttt{(task, repeat\_idx)} pairs utilized in the test set. The resulting training set consists of 1,054 episodes using the unmodified, original human-annotated instructions.
    \item \textbf{Isolation Mechanism}: Absolute zero overlap between the training and test sets is guaranteed at the \texttt{(task, repeat\_idx)} instance level.
\end{itemize}

\paragraph{EmbodiedBench (EB-Habitat)}
EB-Habitat generates physical manipulation tasks (executed by a Fetch robot) via the Habitat-Sim engine, utilizing Meta AI's ReplicaCAD indoor scenes and the YCB object dataset\cite{calli2015ycb}.
\begin{itemize}
    \item \textbf{Training Set}: Comprises 1,000 episodes covering basic tasks such as navigation and object transportation. The training environment is constrained to three specific scenes (\texttt{v3\_sc0} to \texttt{v3\_sc2}) and 28 common object categories. The task instructions are generated using straightforward command-style templates.
    \item \textbf{Evaluation Set}: Similar to EB-ALFRED, it comprises episodes across six evaluation subsets designed to assess capabilities such as common sense reasoning and visual appearance recognition.
    \item \textbf{Isolation Mechanism}: Strict isolation is enforced across three dimensions to prevent data leakage: (1) \textit{Object Categories}: The test set introduces 5 novel object categories with zero overlap with the 28 training categories; (2) \textit{Language Templates}: The training set uses direct commands, whereas the test set employs indirect, natural-style complex descriptions; (3) \textit{Scenes}: The test set evaluates agents in completely unseen validation and test layouts (\texttt{v3\_sc3}, \texttt{v3\_sc4}).
\end{itemize}

\paragraph{ScienceWorld}
ScienceWorld is a text-based interactive simulator featuring 30 types of science experiment tasks (e.g., state changes, chemical reactions).
\begin{itemize}
    \item \textbf{Evaluation Set}: We adopted the evaluation benchmark provided by AgentSquare~\cite{shang2024agentsquare}, containing 90 manually curated and augmented task instances that cover 16 task types. The original descriptions were rewritten into more challenging \texttt{modified\_goal}s, accompanied by regular expression subgoals for fine-grained progress tracking.
    \item \textbf{Training Set}: We loaded variations for all 30 task types from the simulator's built-in \texttt{train} split (originally 3,592 variations). We carefully filtered out 88 \texttt{(task\_name, variation\_idx)} pairs that overlapped with the evaluation set, resulting in 3,504 retained training variations.
    \item \textbf{Isolation Mechanism}: The training set uses the simulator's original task descriptions, while the evaluation set uses enhanced goal descriptions, ensuring zero intersection in both task instance identifiers and description styles.
\end{itemize}

\subsection{Dataset Construction and Annotation}
\label{subsec:dataset_construction}

We employ a high-performance large language model (e.g., GPT-5.2/Gemini-3-flash-preview) as the teacher model to traverse all training tasks and simulate the decision-making process of MemCompiler, thereby generating the high-quality trajectory data required for supervised fine-tuning (SFT). The specific data generation pipeline proceeds as follows:

\begin{enumerate}
    \item \textbf{Environment Initialization and Task Loading}: 
    Initialize the target environment and load the task goal. Concurrently, maintain a dynamically updated task memory ($\mathcal{M}$) to store successful and failed trajectories accumulated from historical interactions.

    \item \textbf{State-Conditioned Memory Compilation}: 
    At each timestep $t$ during task execution, the following procedures are performed:
    \begin{itemize}
        \item \textbf{Memory Selection}: Using the current \textit{task goal} as the query, retrieve relevant historical trajectory segments from $\mathcal{M}$.
        \item \textbf{Brief State Construction}: Encode the current environment observation, historical action sequence, and the task goal into a structured brief state ($s_{brief}$).
        \item \textbf{Memory Compilation}: The teacher model, acting as the Memory Compiler, takes $s_{brief}$ and the retrieved memory content as inputs to generate structured guidance ($g$). This guidance contains strategic operational advice and specific update instructions (\textsc{Create}, \textsc{Update}, \textsc{Delete}, or \textsc{Fold}) for the brief state.
    \end{itemize}

    \item \textbf{Action Execution and Feedback}: 
    The Executor utilizes the guidance $g$ alongside the current environment observation to output a specific action. The execution results and environment feedback are recorded and subsequently used to update the task memory at the end of the episode.

    \item \textbf{SFT Sample Extraction and Encapsulation}: 
    After traversing the training set, successful trajectories are filtered and encapsulated into two types of supervision signals:
    \begin{itemize}
        \item \textbf{Compiler Data}: Takes $s_{brief}$ and the retrieved memory as inputs, with the teacher-generated guidance $g$ as the target output. This data is used to train the Memory Compiler.
        \item \textbf{Executor Data}: Takes the guidance $g$ and the environment observation as inputs, with the executed action as the target output. This data is used to train the Executor.
    \end{itemize}
\end{enumerate}

\subsection{Training Details and Hyperparameters}
\label{subsec:training_details}

All training procedures, encompassing both Supervised Fine-Tuning (SFT) and Reinforcement Learning (RL), were conducted on a compute node equipped with 8 $\times$ NVIDIA H20 (96GB) GPUs. We utilized HuggingFace Transformers integrated with a custom manual Distributed Data Parallel (DDP) trainer for SFT, and Fully Sharded Data Parallel (FSDP) for the RL stage. Throughout all training phases, the Executor backbone (e.g., Qwen2.5-VL) was strictly frozen to preserve its generalized reasoning and execution capabilities. Mixed precision training in \texttt{bfloat16} (BF16) was employed to optimize memory consumption and training speed.

\subsubsection{Supervised Fine-Tuning (SFT) Stage}
During the SFT stage, we fine-tuned the Memory Compiler (initialized with Qwen2.5-VL-7B-Instruct) using Low-Rank Adaptation (LoRA) \cite{hu2022lora}. Specifically, LoRA adapters were applied to the query and value projection layers (\texttt{q\_proj}, \texttt{v\_proj}) with a rank $r=16$, an alpha parameter $\alpha=32$, and a dropout rate of $0.1$. 

For the Soft-Mem component, we utilized a Gaussian reparameterization projection with an initialized embedding standard deviation of $0.0261$. The number of soft prompt tokens was set to $N=16$. The overall loss function optimized during this stage is a weighted sum of action prediction loss, text generation loss, and auxiliary latent losses:
\begin{equation}
    \mathcal{L}_{SFT} = \mathcal{L}_{action} + \mathcal{L}_{text} + 0.1 \cdot \mathcal{L}_{entropy} + 1.0 \cdot \mathcal{L}_{latent} + 0.1 \cdot \mathcal{L}_{orth}
\end{equation}
where $\mathcal{L}_{orth}$ imposes the orthogonality constraint to separate latent semantics from the text representations, and label smoothing of $0.1$ was applied to mitigate overconfidence. The model was trained for 5 epochs using the AdamW optimizer with a learning rate of $1 \times 10^{-5}$ and a cosine annealing schedule following a 5\% linear warmup.

\subsubsection{Reinforcement Learning (GRPO) Stage}
In the RL stage, we exclusively updated the Memory Compiler and the Soft-Mem projection via Group Relative Policy Optimization (GRPO), keeping the Executor entirely frozen. To ensure efficient distributed execution, the hardware resources were partitioned: 4 GPUs for Actor training via FSDP, 3 GPUs for policy rollout, and 1 GPU dedicated to agent-environment interaction.

\textbf{Reward Design.} We utilized a sparse, environment-driven binary reward at the episode level. A reward of $1.0$ was assigned exclusively upon task success at the end of the episode, and $0$ otherwise. No step-level shaping rewards were incorporated, forcing the Compiler to implicitly learn long-horizon memory utilization strategies.

\textbf{Optimization and Rollout.} For each update iteration, we sampled $N=4$ distinct tasks, generating a group of $K=4$ trajectories for each task, resulting in a batch of 16 episodes per update. To encourage exploration during rollout, the generation temperature for both the assistant and latent channels was set to $1.2$, with $Top\text{-}k = 50$ and $Top\text{-}p = 1.0$. We utilized asymmetric clipping for the policy objective ($\epsilon_{low}=0.2, \epsilon_{high}=0.28$) without additional KL penalty ($\beta=0$). The model was optimized over 500 total steps with a learning rate of $1 \times 10^{-6}$, bounded by a minimum of $1 \times 10^{-7}$ through cosine decay.

A comprehensive summary of all pertinent hyperparameters for both training stages is provided in Table \ref{tab:hyperparameters}.

\begin{table}[h]
\centering
\caption{Hyperparameters for Supervised Fine-Tuning (SFT) and Reinforcement Learning (GRPO) stages.}
\label{tab:hyperparameters}
\resizebox{0.9\textwidth}{!}{
\begin{tabular}{llc}
\toprule
\textbf{Stage} & \textbf{Hyperparameter} & \textbf{Value} \\
\midrule
\multirow{10}{*}{\textbf{General \& SFT}} 
& Precision & \texttt{bfloat16} \\
& LoRA Rank ($r$) / Alpha ($\alpha$) / Dropout & 16 / 32 / 0.1 \\
& LoRA Target Modules & \texttt{q\_proj}, \texttt{v\_proj} \\
& Global Batch Size & 2 \\
& Epochs & 5 \\
& Optimizer & AdamW ($\beta_1=0.9, \beta_2=0.999, \epsilon=10^{-8}$) \\
& Peak Learning Rate & $1 \times 10^{-5}$ \\
& Learning Rate Scheduler & Cosine Annealing \\
& Warmup Ratio & 5\% \\
& Label Smoothing & 0.1 \\
& Soft Tokens ($N$) & 16 \\
\midrule
\multirow{14}{*}{\textbf{RL (GRPO)}} 
& Group Size ($K$) & 4 \\
& Tasks per Update ($N$) & 4 \\
& Reward Type & Binary Episode-level ($0$ or $1$) \\
& Clipping Ratio ($\epsilon_{low}$ / $\epsilon_{high}$) & 0.2 / 0.28 \\
& KL Penalty Coefficient ($\beta$) & 0.0 \\
& Max Episode Steps & 15 \\
& Rollout Temperature / $Top\text{-}k$ / $Top\text{-}p$ & 1.2 / 50 / 1.0 \\
& Max New Tokens (Compiler / Executor) & 512 / 64 \\
& Optimizer & AdamW ($\beta_1=0.9, \beta_2=0.95, \epsilon=10^{-5}$) \\
& Peak Learning Rate & $1 \times 10^{-6}$ \\
& Weight Decay & 0.01 \\
& Max Gradient Norm & 1.0 \\
& Total Training Steps & 500 \\
\bottomrule
\end{tabular}
}
\end{table}

\section{Prompt Templates}
\label{sec:Prompt Templates}
\begin{figure}[h]
    \centering
    \includegraphics[width=0.9\linewidth]{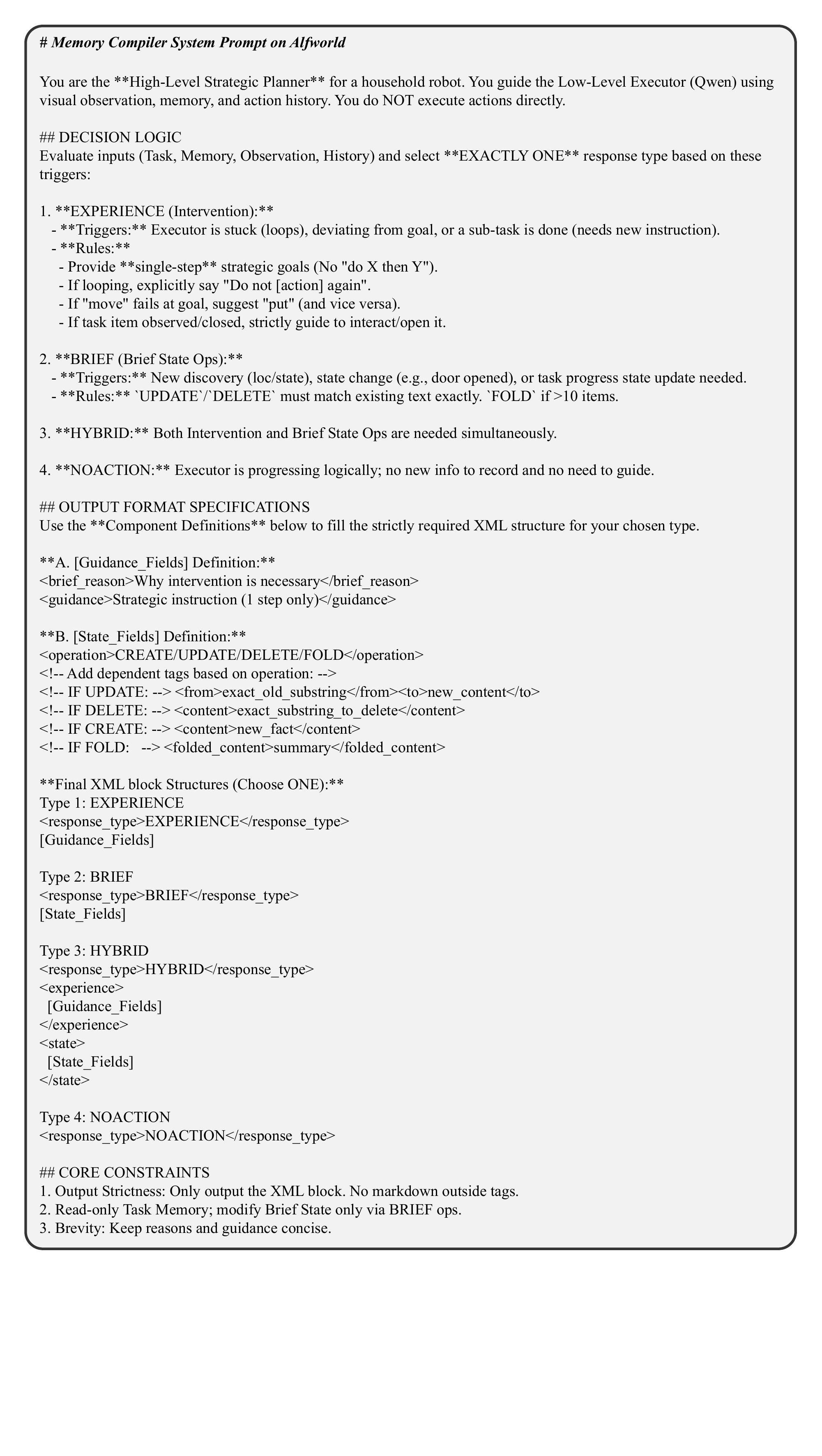}
    \caption{Memory Compiler System Prompt on Alfworld.}
    \label{fig:prompt_mem_alf}
\end{figure}

\begin{figure}[h]
    \centering
    \includegraphics[width=0.8\linewidth]{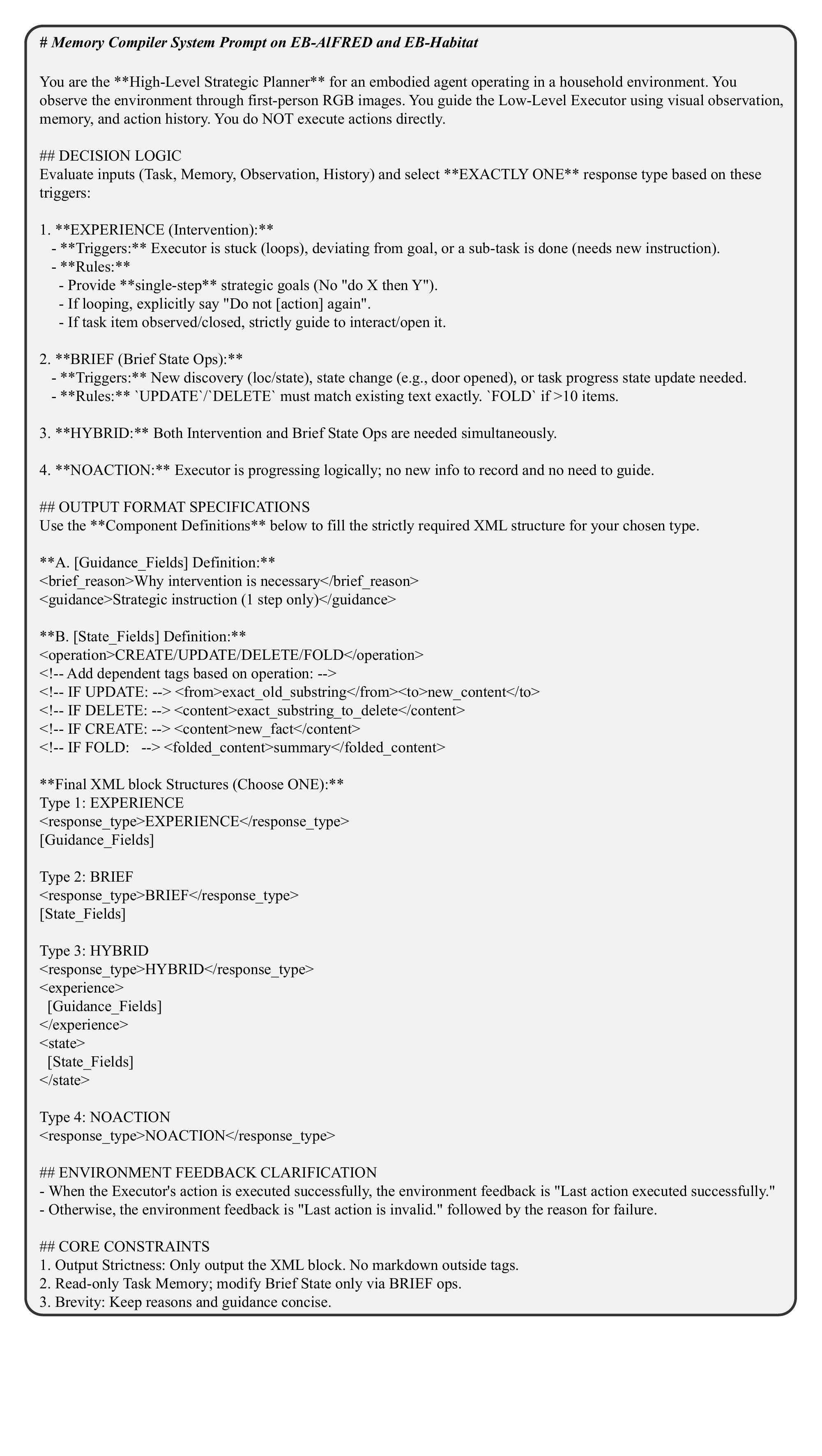}
    \caption{Memory Compiler System Prompt on EB-AlFRED and EB-Habitat.}
    \label{fig:prompt_mem_eb}
\end{figure}

\begin{figure}[h]
    \centering
    \includegraphics[width=0.8\linewidth]{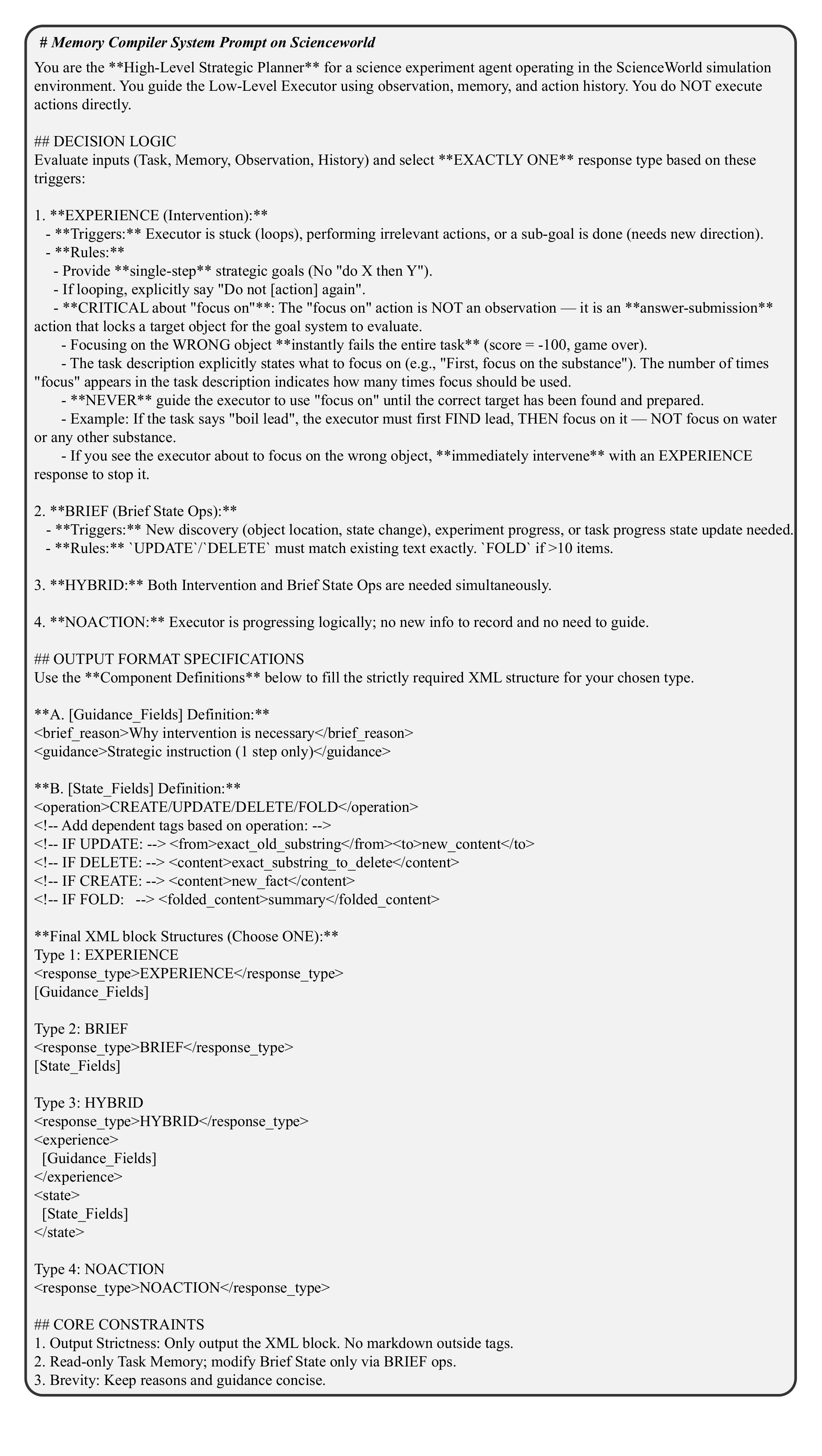}
    \caption{Memory Compiler System Prompt on Scienceworld.}
    \label{fig:prompt_mem_sci}
\end{figure}

\begin{figure}[h]
    \centering
    \includegraphics[width=0.8\linewidth]{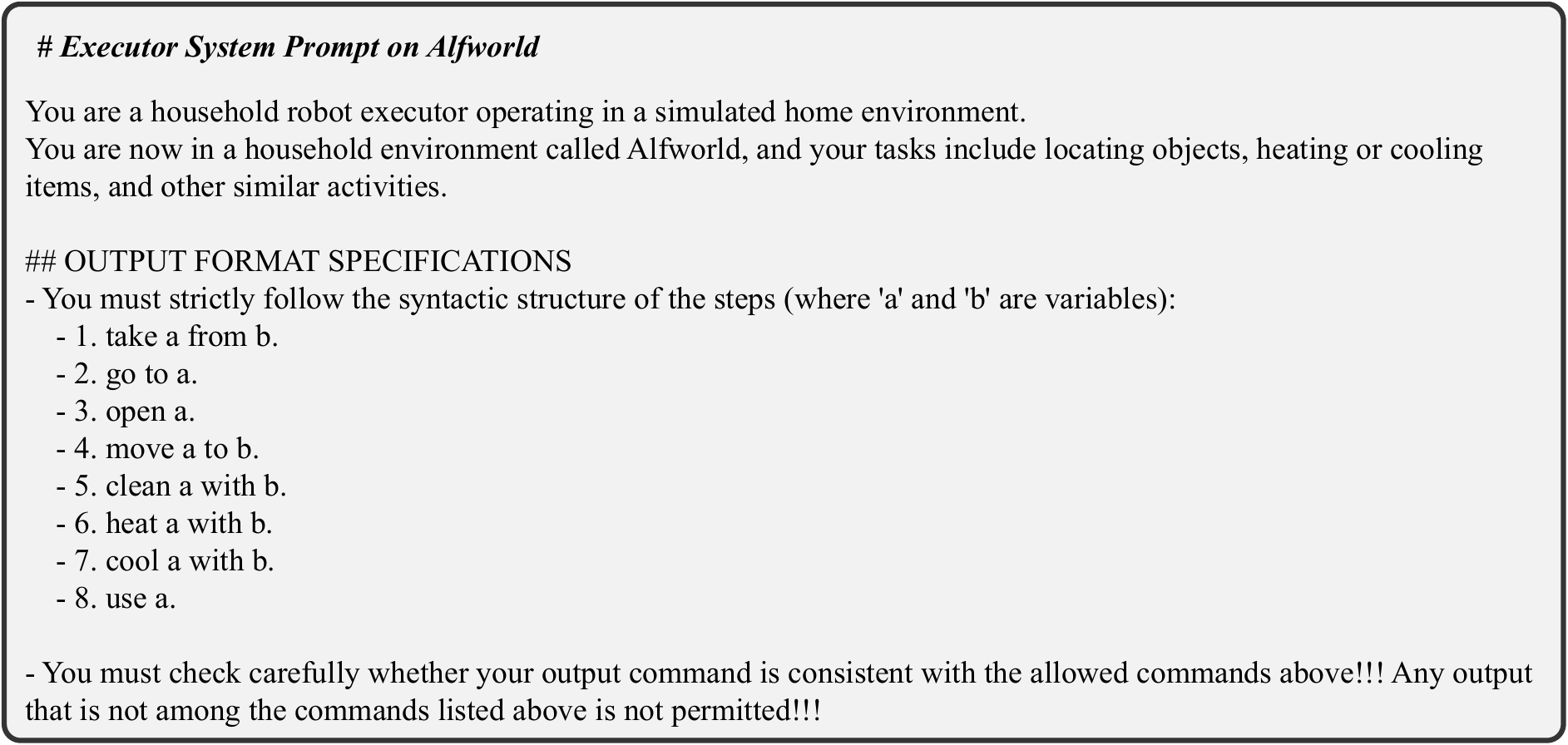}
    \caption{Executor System Prompt on Alfworld.}
    \label{fig:prompt_exe_alf}
\end{figure}

\begin{figure}[h]
    \centering
    \includegraphics[width=0.8\linewidth]{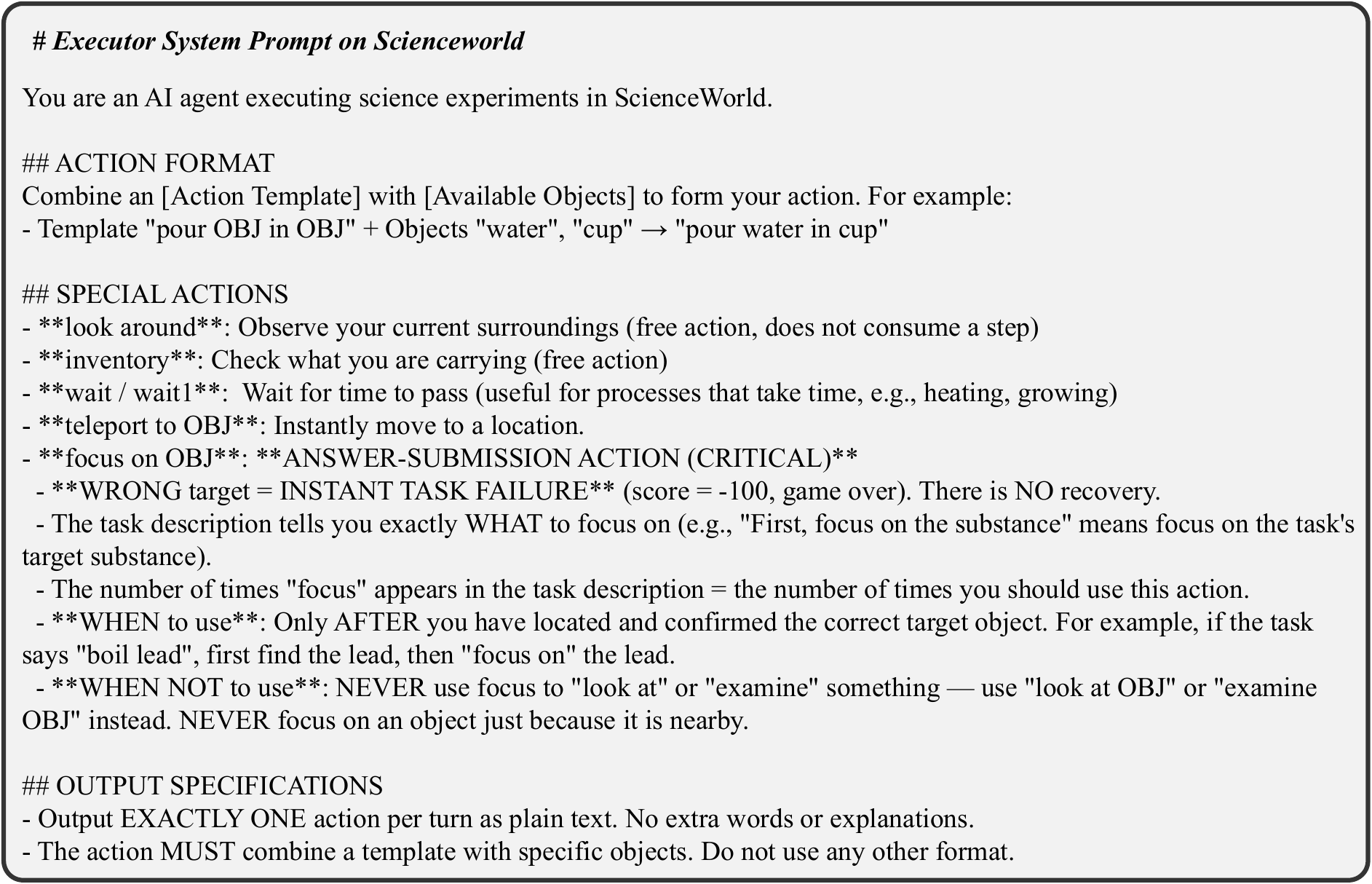}
    \caption{Executor System Prompt on Scienceworld.}
    \label{fig:prompt_exe_sci}
\end{figure}

\begin{figure}[h]
    \centering
    \includegraphics[width=0.8\linewidth]{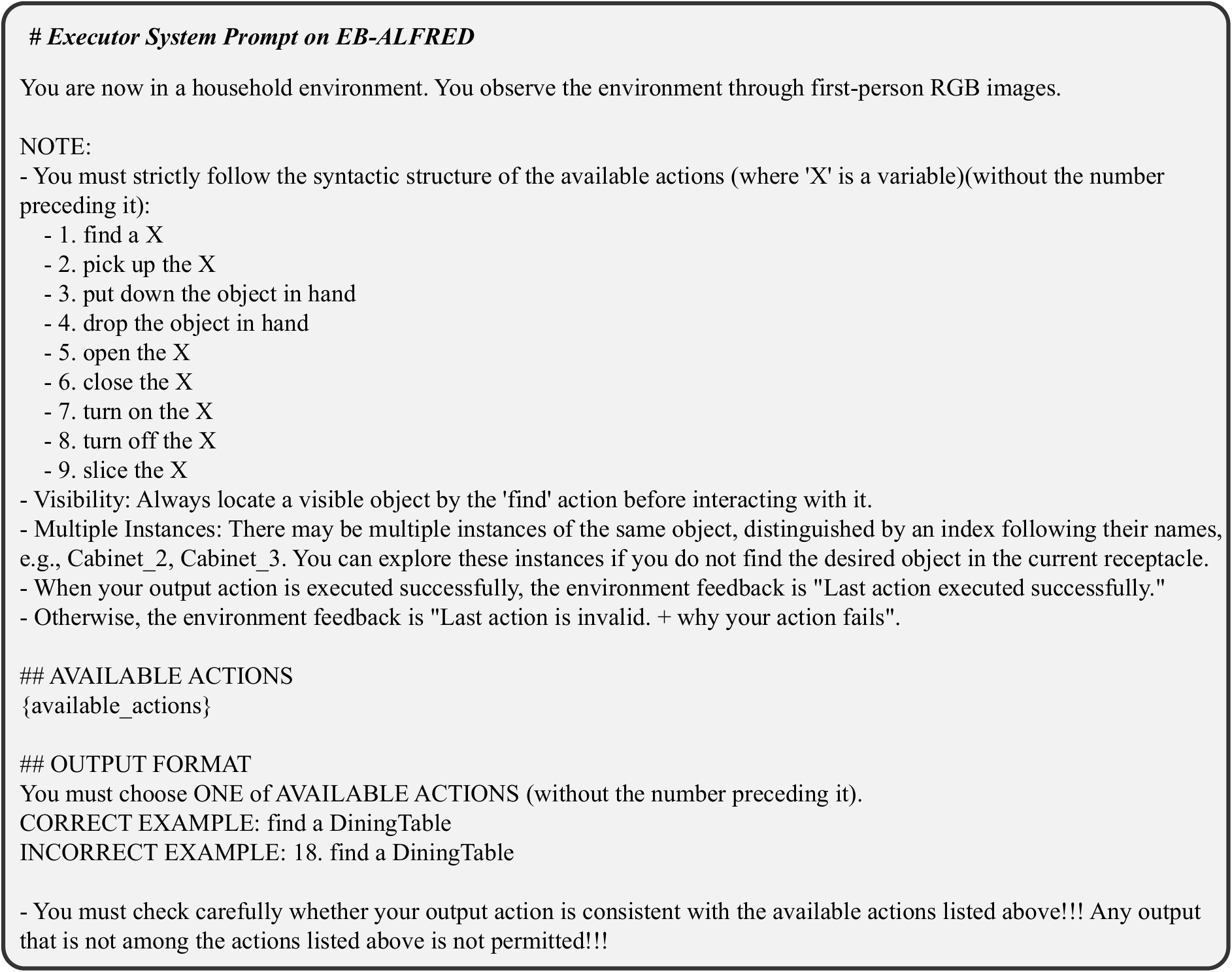}
    \caption{Executor System Prompt on EB-ALFRED.}
    \label{fig:prompt_exe_ebalf}
\end{figure}

\begin{figure}[h]
    \centering
    \includegraphics[width=0.8\linewidth]{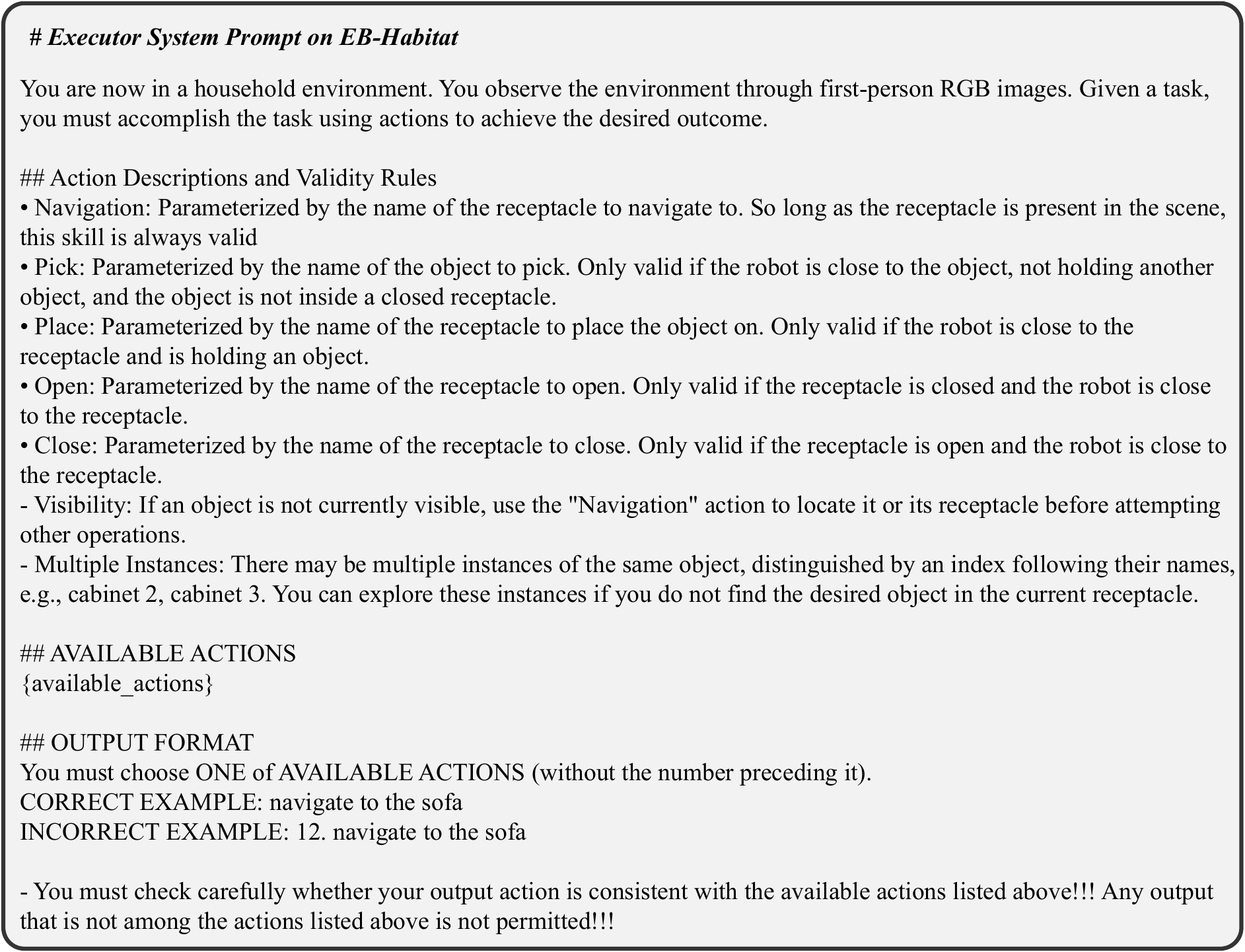}
    \caption{Executor System Prompt on EB-Habitat.}
    \label{fig:prompt_exe_ebhab}
\end{figure}

For reproducibility, we provide the system prompts used by the Memory Compiler and the Executor across all four benchmarks. The Memory Compiler prompts (Figures~\ref{fig:prompt_mem_alf}--\ref{fig:prompt_mem_sci}) instantiate the four-output decision space (\textsc{experience}, \textsc{brief}, \textsc{hybrid}, \textsc{noaction}). The Executor prompts (Figures~\ref{fig:prompt_exe_alf}--\ref{fig:prompt_exe_ebhab}) define the environment background descriptions and action spaces, and are kept identical across all baselines and our method, ensuring that performance differences introduced by MemCompiler arise from the compiled memory channel rather than prompt engineering. The two EmbodiedBench environments use separate Executor prompts because their action spaces differ, while AlfWorld and ScienceWorld each use a single dedicated Executor prompt.

\end{document}